\documentclass{article}
\usepackage{a4wide}
\usepackage{amsthm}
\usepackage{amsfonts}
\usepackage{latexsym}
\usepackage{amssymb}
\usepackage{amsmath}
\usepackage{epic}
\usepackage{eepic}
\usepackage{array}
\usepackage{xspace}
\usepackage{varioref}
\renewcommand{\a}{\mbox{*}}

\newcommand{\A}{\Sigma}
\newcommand{\symb}[1]{\mbox{\sffamily #1}}

\newcommand{\auto}{\mathcal{A}}
\newcommand{\rules}{\Delta}
\newcommand{\fauto}{\mathit{FA}}

\newcommand{\Moverel}[2][\auto]{\mbox{$\:\scriptstyle|\negthickspace\frac{\scriptstyle\,#2}{\makebox[2.5ex]{$\scriptscriptstyle{#1}$}}\:$}\xspace}
\newcommand{\moverel}{\Moverel{}}

\newcommand{\derivrel}{\,\raisebox{3pt}{$\underset{G}{\raisebox{-3pt}[-3pt][-3pt]{$\longrightarrow$}}$}\,}

\newcommand{\foo}{string\xspace}
\newcommand{\Foo}{String\xspace}
\newcommand{\inst}{\lhd}
\newcommand{\steps}[2]{\mathrm{Steps}^{#1}_{#2}}
\newcommand{\basis}{\medskip\par\noindent\textit{Basis}\quad}
\newcommand{\induction}{\medskip\par\noindent\textit{Induction}\quad}
\newcommand{\there}{\medskip\par\noindent$\Longrightarrow$:\quad}
\newcommand{\back}{\medskip\par\noindent$\Longleftarrow$:\quad}

\newcommand{\RE}{\ensuremath{\mathit{RE}}\xspace}

\newcommand{\reglang}[1]{\ensuremath{[\![{#1}]\!]}\xspace}
\newcommand{\viter}[1]{^{\textstyle\ast_{#1}}}
\newcommand{\anytree}{\Theta}

\theoremstyle{definition}
\newtheorem{definition}{Definition}
\newtheorem{example}{Example}

\theoremstyle{remark}

\theoremstyle{plain}

\newtheorem{proposition}{Proposition}
\newtheorem{corollary}{Corollary}
\newtheorem{theorem}{Theorem}
\newtheorem{lemma}{Lemma}

\newcounter{mycnt}
\newcommand*{\mycounter}[1]{\renewcommand*{\themycnt}{#1}\setcounter{mycnt}{0}}
\newcommand{\cstep}{\refstepcounter{mycnt}\themycnt}

\newcommand{\bcategory}[4]{#1 [\textbf{#2}]: #3---\textit{#4}}
\newcommand{\scategory}[3]{#1: #2---\textit{#3}}
\newcommand{\terms}[1]{\medskip\par\noindent General Terms: #1}
\newcommand{\keywords}[1]{\medskip\par\noindent Additional Key Words and
Phrases: #1}

\begin{document}
\title{Using Tree Automata and Regular Expressions\\
to Manipulate Hierarchically Structured Data}
\author{Nikita Schmidt and Ahmed Patel\\University College Dublin\thanks{
The authors are with the Computer Networks and Distributed Systems Research
Group, Department of Computer Science, University College Dublin, Belfield,
Dublin 4, Ireland.  E-mail: cetus@cnds.ucd.ie, apatel@cnds.ucd.ie.  Tel.:
+353-1-7162476.}
}
\date{}
\maketitle
%\renewcommand{\baselinestretch}{1.7}
%\vspace{-0.5cm}
%\small\begin{center}\bf Abstract\end{center}
\begin{abstract}
Information, stored or transmitted in digital form, is often structured.
Individual data records are usually represented as hierarchies of their
elements.  Together, records form larger structures.  Information processing
applications have to take account of this structuring, which assigns different
semantics to different data elements or records.  Big variety of structural
schemata in use today often requires much flexibility from applications---for
example, to process information coming from different sources.  To ensure
application interoperability, translators are needed that can convert one
structure into another.

This paper puts forward a formal data model aimed at supporting hierarchical
data processing in a simple and flexible way.  The model is based on and
extends results of two classical theories, studying finite string and tree
automata.  The concept of finite automata and regular languages is applied to
the case of arbitrarily structured tree-like hierarchical data records,
represented as ``structured strings.'' These automata are compared with
classical string and tree automata; the model is shown to be a superset of the
classical models.  Regular grammars and expressions over structured strings
are introduced.

Regular expression matching and substitution has been widely used for
efficient unstructured text processing; the model described here brings the
power of this proven technique to applications that deal with information
trees.  A simple generic alternative is offered to replace today's specialised
ad-hoc approaches.  The model unifies structural and content transformations,
providing applications with a single data type.  An example scenario of how to
build applications based on this theory is discussed.  Further research
directions are outlined.
%Information processing applications have to take account of information
%structure along with its content.  Building of such applications requires a
%data model that can represent and operate with structured documents.  This
%paper puts forward a flexible formal data model, based on string and tree
%automata theories.  The concept of tree regular expressions is extended so as
%to be applicable to general purpose processing of information trees.  The
%model unifies structural and content transformations, providing applications
%with a single data type.  A sample application scenario is discussed.
%\medskip

%\textbf{Keywords:}
%Information structure, Hierarchy, Tree automaton, Regular expression, 
%Data model
%\end{abstract}

\bigskip\noindent
Categories and subject descriptors:
\scategory{E.1}{Data Structures}{Trees};
\bcategory{F.4.3}{Mathematical Logic and Formal Languages}{Formal
Languages}{Classes defined by grammars or automata};
\bcategory{I.7.2}{Document and Text Processing}{Document Preparation}{Markup
languages};
\bcategory{H.3.3}{Information Storage and Retrieval}{Information Search and
Retrieval}{Query formulation}

\terms{Theory, Languages} 

%\begin{bottomstuff}
%Authors' address:
%Computer Networks and Distributed Systems Research Group,
%Department of Computer Science, University College Dublin, Belfield, Dublin 4,
%Ireland.  E-mail: cetus@cnds.ucd.ie, apatel@cnds.ucd.ie.  Tel: +353-1-7162476.
%\end{bottomstuff}
%\maketitle
\end{abstract}
%\normalsize
%\addtolength{\textheight}{-60pt}

\section{Introduction}

Information processing has always faced the need to take into account the
structure of the data being processed.	Structuring of information plays an
important role in fostering automated, computerised data capture, storage,
search, retrieval, and modification.  For example, an unstructured
bibliographic reference like `Bourbaki,~N. Lie Groups and Lie Algebras'
requires either a human assistance or the use of heuristics to determine
whether Lie is the name of the author or a part of the title.  On the other
hand, dividing that reference in two parts---`author' and `title'---from the
very beginning would have solved this problem.  The `author' part can be
further sub-structured into `last name', `first initial' and so on.

The structure of data records depends on the kind of information the records
carry---for example, flights schedule, asset list, book, e-mail message, and
so on.	It is often necessary to convert data between different structured
representations, along with more basic tasks such as retrieval of certain
components of a record or addition of a new component.

Hierarchical (tree-like) way of organising information is very popular and
convenient, because it allows aggregation of details at different granularity
levels.  Most of the data structures used today either are already
hierarchical, or can be expressed using an information tree~\cite{lotusdm}.
XML~\cite{XML-spec} is being increasingly used as the standard language for
representing hierarchically structured data.  Areas where structured
information is actively utilised include:
\begin{itemize}
\item text processing (markup languages);
\item information retrieval (document processing, query adaptation);
\item compilers (syntax trees);
\item library automation (bibliographic records);
\item a wide variety of industrial applications.
\end{itemize}

%\addtolength{\topmargin}{60pt}

This paper puts forward a simple and flexible formal data model for
manipulating structured information.  This model is built on top of a
combination of results from finite automata and tree automata theories.  The
concept of automata and regular languages is applied to the case of
arbitrarily structured tree-like hierarchical data records, represented as
``structured strings.''  The paper compares these automata with classical
string and tree automata, showing that the theory presented here is a superset
of the classical models: everything that can be done with finite string or
tree automata can also be done in our model.

The data manipulation model suggested here is based around tree regular
expressions, their matching and substitution.  Regular expressions are widely
used in non-structured text processing, serving as a core model in many text
processing applications~\cite{sed-awk}.  They are good for selecting fragments
that match certain patterns in certain contexts.  For example, regular
expression `\verb|Figure +([0-9]+)|' selects all figure numbers in a text
document (more precisely, it selects all decimal numbers that follow the word
`Figure', separated by at least one space).

This paper brings the power of regular expressions to serve hierarchical data
manipulation.  The notion of regular expressions is extended to information
trees in a way that matches most applications' needs.  This offers a single,
simple generic solution to many problems where different incompatible ad-hoc
approaches have been used before.  Although the model is described from a
theoretical standpoint, its practical applications are considered, drawing on
the experience of using regular expression matching techniques on plain text
documents.  There is room for further research in terms of both theory and
applications. 

The rest of the paper is organised as follows.	The next section surveys
related works and discusses approaches taken previously.
Section~\ref{sect:inf_intr} introduces the data model proposed for
representation of information trees, and Section~\ref{sect:model_def} provides
its formal definition.  Then, Section~\ref{sect:automata} defines finite tree
automata that operate on information trees, and discusses properties of such
automata.  Section~\ref{sect:grammars} introduces regular grammars and
expressions and shows their equivalence to the finite tree automata.
Section~\ref{sect:app_scenarios} presents an example application scenario to
illustrate how the described theory can be used in practice.
Section~\ref{sect:conclusions} concludes the paper and outlines further
research directions.

\section{Related works}

\subsection{Murata's forest algebra and tree automata theory}

The most recent research on formal tree-structured data models belongs to
Makoto Murata, who applied and extended the theory of tree
automata~\cite{gecseg-steinby:automata} to the problem of transformation of
SGML/XML documents
\cite{murata:DTDtrans,murata:data_model,murata:hedge-automata}.
In~\cite{murata:trans_math} Murata offers a hierarchical data model based on
tree automata theory and the work of Podelski on pointed trees
\cite{podelski}.  However, Murata uses tree automata for purely
representational purposes, as a means to formally define XML schema in his
model, rather than as a main processing tool.

The theory of tree automata~\cite{gecseg-steinby:automata} studies classes of
trees, called \emph{languages}, in a way similar to the formal language
theory.  A tree language is defined by its syntax, which can, by analogy with
the language theory, be described by either a \emph{tree grammar} or a
\emph{tree automaton}.	Murata shows the close relation between SGML or XML
document syntax (usually referred to as \emph{Document Type Definition}, or
\emph{DTD}) and the syntax of its tree representation.  Because the classical
theory only deals with tree languages with limited branching (the maximum
number of branches that any node of any tree of a language may have is a
constant determined by the syntax of that language), Murata had to extend the
theory to handle unlimited branching, necessary to represent marked-up
documents.

In~\cite{murata:data_model}, Murata suggests a data model for transformations
of hierarchical documents.  Although primarily intended to serve the SGML/XML
community, the model is a generally applicable forest-based model.  Murata's
extended tree automata are used for schema representation, parallel to DTDs in
SGML and XML.  The core of the model is a forest algebra, containing fourteen
operators for selecting and manipulating document forests.  This algebra can
select document fragments based on \emph{patterns} (conditions on descendent
nodes) and \emph{contextual conditions} (conditions on non-descendent nodes,
such as ancestors, siblings, etc.).  One of the strong points of Murata's data
model is that during transformation of documents, syntactical changes are
tracked in parallel: operators apply not only to the document being processed,
but also to its schema, so that at any stage in a transformation process, the
schemata of the intermediate results are known.

\subsection{XPath}

A different approach to transformations of hierarchically organised documents
is taken by the World Wide Web Consortium in their XPath~\cite{XPath} and
XSLT~\cite{xslt} specifications.  Although designed specifically for the XML
document representation format, these recommendations are applicable to a wide
variety of other types of documents that can be reasonably translated into
XML.  The XML Path Language (XPath) provides a common syntax and semantics for
addressing parts of an XML document---functionality used by other
specifications, such as XSL Transformations (XSLT).  XPath also has facilities
for manipulating strings, numbers, and Booleans, which support its primary
purpose.

XPath's model of an XML document is that of an ordered tree of nodes, where
nodes can be of seven types.  The multitude of types is needed to support
various XML features, such as namespaces, attributes, and so on.  XPath can
operate on documents that come with or without a Document Type Definition
(DTD).  A DTD, when supplied, unlocks some functionality of the XPath
processor, such as the ability to find unique IDs of document elements, or to
use default attribute values.

XPath is an expression-based language.  Expressions evaluate to yield objects
of four basic types: node-set, Boolean, number (floating-point), and string.
Expressions consist of string and numeric constants, variable references,
unary and binary operators, function calls, and special tokens.  The
specification defines core function library that all XPath implementations
must support.  The library contains 7 node-set functions, 10 string, 5
Boolean, and 5 numeric functions.  Some of the XPath operators and functions,
such as \texttt{+}, \texttt{-}, \texttt{floor}, \texttt{string-length},
\texttt{concat}, are of general purpose nature and are typical of traditional
programming languages.

\section{Hierarchical data model: An informal introduction}
\label{sect:inf_intr}

The basic data model that we use in this paper was originally introduced
in~\cite{lotusdm}.  It was also shown there that all popular information
structuring methods can be realised using tree-like structures and expressed
by this model.	We give here its brief description.

\begin{figure}[htb]
\begin{center}
\begin{tabular}{c@{\quad\quad}c@{\quad\quad}c}
\hspace{10em}		&
\input{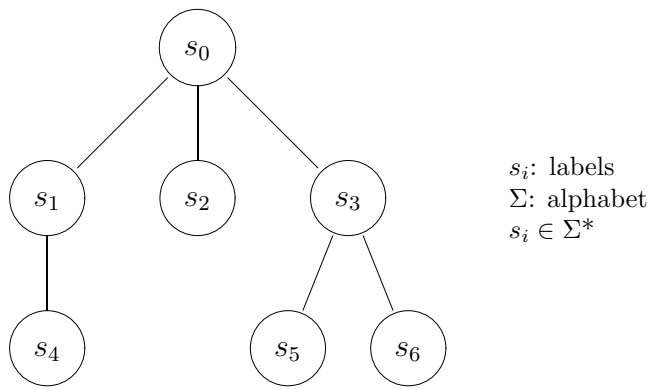}	&
\parbox[b][5cm][c]{10em}{$s_i$: labels\\$\A$: alphabet\\$s_i\in\A\a$}
\end{tabular}
\end{center}
\caption{Labelled tree over alphabet $\A$}
\label{fig:lbtree}
\end{figure}

Informally, in the proposed model a document is represented as a finite
ordered labelled tree.	Each node of the tree is associated with a
\emph{label}: a string over an alphabet (see Figure~\ref{fig:lbtree}).	In the
traditional terminology, the labels of leaf, or terminal, nodes of a document
are called \emph{data elements}.  They carry the ``actual'' content of the
document.  The labels of internal (non-leaf) nodes are referred to as
\emph{tags}, whose purpose is to describe those data elements.	When speaking
of tags and data elements, we shall often make no distinction between nodes
and their labels.

Tags form the structure of a document, specifying the semantics of their
underlying sub-trees (and, ultimately, the data elements).  The sequence of
tags from the root to a data element is called a \emph{tag path} and fully
identifies the properties and interpretation of that element.  Sometimes tag
paths are used as keys to extract data elements from documents.

This model has the following properties:
\begin{itemize}
\item Unlimited branching: although all trees are finite, the number of
children a node may have can be arbitrarily large.  By contrast, in the
tree automata theory branching is limited and is determined by the tree's
ranked alphabet.
\item Unlimited number of possible labels: labels are selected from an
infinite set of strings over a finite alphabet.
\item Tags and data elements, which are traditionally regarded as belonging to
different domains, are built here uniformly from the same alphabet.  This
allows the use of the same operators and mechanisms for both leaves and
internal nodes.  Information contained in data elements and tags is freely
interchangeable.
\item This model can be used for string manipulations, whereby strings are
represented as single-node trees.
\end{itemize}
The last property suggests that traditional string operators could probably be
extended to the tree case.  In other words, there is an opportunity to design
a good tree algebra in such a way that restricting it to single node trees
would result in a meaningful and convenient string algebra.

\section{Model definition}
\label{sect:model_def}

This section presents a formal definition of the model described above.  We
start it by introducing the terminology used in the rest of the paper.
\begin{itemize}
\item $\A$ is a finite set of symbols, called \emph{alphabet}.	For notational
convenience, we assume that $\A$ does not contain angle brackets and slash:
$\langle,\rangle,/ \not\in \A$.
\item $\A\a$ is the free monoid on $\A$, or the set of all strings over $\A$
together with the concatenation operator.
\item $\varepsilon$ is the empty string; $\varepsilon\in\A\a$.
\end{itemize}
All the examples given below are based on the alphabet of Latin letters.

\subsection{\Foo trees}

We introduce \emph{\foo trees} over an alphabet $\A$ as strings with angle
brackets such that (a) brackets match pairwise, and (b) each whole tree is
enclosed in a pair of brackets, for example:
\symb{$\langle$ab$\langle$cde$\rangle$f$\langle$g$\langle$hi$\rangle\rangle
\rangle$}.
Visually similar to the tree representations found in classical
literature~\cite{knuth} and in recent research~\cite{murata:data_model}, \foo
trees bear one significant difference.	Traditionally, each symbol marked a
separate node; brackets contained all the children of the node marked by the
symbol immediately preceding the opening bracket.  In our model, each node is
labelled by a \emph{sequence} of symbols, enclosed in a pair of brackets.

The set $T(\A)$ of \foo trees over $\A$ is defined as the minimum 
subset of $(\A\cup\{\langle,\rangle\})\a$ such that:
\begin{enumerate}
\item\label{item:null} $\langle\rangle \in T(\A)$ (the null tree)
\item\label{item:alph} $\langle a\rangle \in T(\A)$ for any $a \in \A$
\item\label{item:conc} $\langle xy\rangle \in T(\A)$ for any $\langle
	x\rangle,\langle y\rangle \in T(\A)$ (concatenation)
\item\label{item:encp} $\langle t\rangle \in T(\A)$ for any $t \in T(\A)$
	(encapsulation).
\end{enumerate}
It follows immediately from this definition that $T(\A)$ forms a monoid with
respect to concatenation (from \ref{item:null} and \ref{item:conc}) and that
all strings from $\A\a$, enclosed in a pair of angle brackets, are contained
in $T(\A)$.

Because concatenation in $T(\A)$ is defined differently from the usual string
concatenation in $(\A\cup\{\langle,\rangle\})\a$, it will be denoted as a
centered dot ($\cdot$).  For example, if $u,v \in T(\A)$, then $u\cdot v$ is
their tree concatenation and $uv$ is their string concatenation.  We could, of
course, discard the outermost pair of angle brackets from trees in $T(\A)$:
they are present in all trees anyway.  This would eliminate the difference
between concatenations of trees and strings, making $T(\A)$ a sub-monoid of
$(\A\cup\{\langle,\rangle\})\a$.  However, this would also complicate automata
and grammars on \foo trees.

We shall often encounter simple cases of trees, consisting of just one label,
such as $\langle$\symb{cetus}$\rangle$, and their subset, single-symbol trees,
such as $\langle$\symb{a}$\rangle$.  The following notation will be used:
\begin{itemize}
\item $\langle\A\rangle$ is the set of all single-symbol trees from $T(\A)$:
	$\{\langle a\rangle\:|\:a\in\A\}$;
\item $\langle\A\a\rangle$ is, by analogy, the set of all single-string trees:
	$\{\langle s\rangle\:|\:s\in\A\a\}$.
\end{itemize}

A matching pair of angle brackets can be thought of as a unary operator, which
we called encapsulation (denoted as $\langle\rangle$).	Note that
encapsulation and concatenation operators are free from relations (apart from
the associativity of concatenation), so they freely generate $T(\A)$ from
$\langle\A\rangle$.  Thus, $T(\A)$ can be called a \emph{free monoid with
unary operator} on $\langle\A\rangle$.

Note that there are two different structures denoted by $T(\A)$: strings over
$\A \cup \{\langle, \rangle\}$ and trees over $\A$.  Strings possess just one
binary operator, concatenation, which has no symbol.  Trees have two
operators, concatenation and encapsulation, denoted by $\cdot$ and
$\langle\rangle$.  In the rest of the paper, we shall be using this dual
notation, where an element of $T(\A)$ can be interpreted as a string or a
tree depending on the context.	The context is uniquely identified by the
operator signs used.  In the text, elements of $T(\A)$ will always be called
trees, to help distinguish them from arbitrary strings from the larger set
$(\A \cup \{\langle, \rangle\})\a$.

\subsection{Reduced \foo trees}

The above definition of \foo trees is, however, too broad for the informal data
model described in Section~\ref{sect:inf_intr}.  Let us consider the
correspondence between the two.

A tree from $T(\A)$ can be uniquely represented as $\langle s_0 t_1 s_1 t_2
s_2 \cdots t_n s_n\rangle$, where $n\ge 0$, $s_i \in \A\a$, and $t_i \in
T(\A)$.  The root label of this tree is $s_0s_1\cdots s_n$, and the children
of the root node are the trees $t_1, t_2, \dots, t_n$.

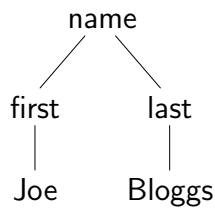
\begin{figure}[htb]
\centering
\setlength{\unitlength}{0.00087489in}
\begingroup\makeatletter\ifx\SetFigFont\undefined%
\gdef\SetFigFont#1#2#3#4#5{%
  \reset@font\fontsize{#1}{#2pt}%
  \fontfamily{#3}\fontseries{#4}\fontshape{#5}%
  \selectfont}%
\fi\endgroup%
{\renewcommand{\dashlinestretch}{30}
\begin{picture}(1264,1209)(0,-10)
\path(462,1044)(192,729)
\path(646,1044)(916,729)
\path(162,504)(162,234)
\path(972,504)(972,234)
\put(567,1089){\makebox(0,0)[b]{\smash{{{\SetFigFont{12}{14.4}{\sfdefault}{\mddefault}{\updefault}name}}}}}
\put(162,549){\makebox(0,0)[b]{\smash{{{\SetFigFont{12}{14.4}{\sfdefault}{\mddefault}{\updefault}first}}}}}
\put(972,549){\makebox(0,0)[b]{\smash{{{\SetFigFont{12}{14.4}{\sfdefault}{\mddefault}{\updefault}last}}}}}
\put(162,54){\makebox(0,0)[b]{\smash{{{\SetFigFont{12}{14.4}{\sfdefault}{\mddefault}{\updefault}Joe}}}}}
\put(972,54){\makebox(0,0)[b]{\smash{{{\SetFigFont{12}{14.4}{\sfdefault}{\mddefault}{\updefault}Bloggs}}}}}
\end{picture}
}
\caption{A tree example}
\label{fig:same_tree}
\end{figure}

It is easily noticeable that the same tree in the informal model can
correspond to different trees in the formal model.  For example, the following
are two different versions of the tree depicted in Figure~\ref{fig:same_tree}:
\symb{$\langle$name$\langle$first$\langle$Joe$\rangle \rangle
	\langle$last$\langle$Bloggs$\rangle \rangle \rangle$},
\symb{$\langle$na$\langle$fir$\langle$Joe$\rangle$st$\rangle$m$\langle
	\langle$Bloggs$\rangle$last$\rangle$e$\rangle$}.

However, a one-to-one correspondence can be achieved by the following
commutativity relation in $T(\A)$:
\begin{displaymath}
\langle a \rangle \cdot \langle t\rangle = \langle t \rangle \cdot \langle
a\rangle\ \mbox{\quad for any}\ a\in \A,\: t\in T(\A).
\end{displaymath}
This relation defines the monoid of \emph{reduced trees}.  This monoid is an
exact match for our informal data model, because the ordering of label symbols
in relation with sub-trees no longer matters: all symbols can be collected in
one part of the label and all sub-trees can be gathered in the other part.

Reduced trees can be constructed as the image of the following map $r:
T(\A)\rightarrow T(\A)$, recursively defined as follows:
\begin{displaymath}
r(u_0\cdot\langle t_1\rangle\cdot u_1\cdot\langle t_2\rangle\cdot u_2\cdot
\ldots \cdot\langle t_n\rangle\cdot u_n) =
u_0\cdot u_1\cdot \ldots \cdot u_n\cdot
\langle r(t_1)\rangle\cdot\langle r(t_2)\rangle\cdot
\ldots \cdot\langle r(t_n)\rangle,
\end{displaymath}
where $n \ge 0$, $u_i \in \langle\A\a\rangle$, and $t_i \in T(\A)$.  This can
also be written in string notation:
\begin{displaymath}
r(\langle s_0 t_1 s_1 t_2 s_2 \cdots t_n s_n\rangle) =
\langle s_0s_1 \cdots s_n r(t_1) r(t_2) \cdots r(t_n)\rangle,
\end{displaymath}
where $n \ge 0$, $s_i \in \A\a$, and $t_i \in T(\A)$.  The binary operator
(concatenation) in Im\,$r$ is naturally defined as
\begin{displaymath}
r(u)\cdot r(v) = r(u\cdot v) .
\end{displaymath}
This operator is well-defined (that is, the result does not depend on the
choice of $u$ and $v$).  Indeed, let us consider $u,\,u',\, v,\,v'$ such that
$r(u) = r(u')$ and $r(v) = r(v')$.  Then, applying the definition of $r$ to
$r(u\cdot v)$ and $r(u'\cdot v')$, we get $r(u\cdot v) = r(u)\cdot r(v) =
r(u')\cdot r(v') = r(u'\cdot v')$.  The associativity of concatenation in Im
$r$ follows immediately from the associativity in $T(\A)$.  Thus, $r(T(\A))$
is a monoid and $r$ is a homomorphism from $T(\A)$ onto $r(T(\A))$.  Note that
despite $r(T(\A))$ being a subset of $T(\A)$, it is not a sub-monoid.

Although reduced trees do provide a better match for actual real-life
hierarchical data structures, normal \foo trees can in fact be more useful
because they give more flexibility in data manipulation.  An actual
transformation engine can easily convert from reduced trees to normal trees
and back if it chooses to work with normal trees internally.

\section{Finite automata}
\label{sect:automata}

The notion of finite automata comes from different branches of computer
science.  A finite automaton is a machine that has a finite set of states, can
accept input from a finite set of input symbols, and changes its state when
input is applied.  The new state depends on the current state and the input
symbol.

In the formal language theory finite automata are used to describe sets of
strings called \emph{regular languages}.  A string is \emph{accepted} by an
automaton if, having consumed all the symbols of the string one by one, the
automaton ends up in a predefined final state.	A set of all strings accepted
by an automaton is called a regular set (or regular language).	The automaton
is said to recognise this language.

Similarly, in the tree automata theory, tree-like structures are operated on
by automata which take symbols in tree nodes as inputs.  There are two classes
of tree automata.  A ``bottom-up'' automaton starts at the leaves and moves
upwards, while a ``top-down'' automaton descends from the root of the tree.
Languages recognised by tree automata constitute the class of \emph{regular
tree languages}.

Let us now introduce finite automata that operate on \foo trees in the
bottom-up manner.  The following definition essentially presents a mixture of
the corresponding notions from the formal language theory and tree automata
theory.

\begin{definition}
A Non-deterministic Finite \Foo Tree Automaton (NFSTA) over $\A$ is a tuple
$\auto = (\A, Q, Q_f, q_0, \rules)$, where
\begin{itemize}
\item $Q \supseteq \A$ is a finite set, called \emph{set of states}, or
	\emph{state set};
\item $Q_f \subseteq Q$ is a set of final states;
\item $q_0 \in Q$ is the initial state;
\item $\rules$ is a set of transition rules of the form $(q_1, q_2) \rightarrow
q_3$, where $q_1,q_2,q_3 \in Q$.
\end{itemize}
$\rules$ can also be thought of as a subset of $Q^3$, however interpreting it
as a set of rules of the above form is more intuitive.
\end{definition}

\begin{definition}
A Deterministic FSTA (DFSTA) is an NFSTA whose $\rules$ contains at most one
rule for each left hand side $(q_1, q_2)$.  In a DFSTA, $\rules$ can also be
considered as a partial function $\rules: Q\times Q \rightarrow Q$.  A DFSTA
whose $\rules$-function is defined on all $Q\times Q$ is called
\emph{complete}.
\end{definition}

The operation of a deterministic FSTA can be illustrated on our informal data
model, introduced in Section~\ref{sect:inf_intr} above, as follows.  The
automaton starts at the leaves.  Each leaf is processed like traditional
finite automata do.  The automaton starts in the state $q_0$ and takes the
first symbol from the leaf's label.  Because that symbol belongs to $\A$, it
also belongs to $Q$.  The automaton finds the rule in $\rules$ whose left-hand
side matches the state and the input symbol.  The right-hand side of that rule
becomes the new state.	Next input symbol is then taken from the label and the
process continues.  If at some stage no rule can be applied, the run is
considered unsuccessful: the tree is not accepted.

When a leaf's label is successfully processed, that leaf is cut off.  The
resulting state of the automaton is inserted into the leaf's parent node label
(the exact place in the label will be discussed later).  As soon as all
children of a node have been processed, the node itself becomes a leaf, and
the automaton runs again.  Note that the definition above allows an automaton
to accept its own state as an input.  If, after processing the root label, the
automaton finishes in one of the ``final states'' ($Q_f$), its run is
considered successful.

A non-deterministic FSTA is different from a DFSTA in that it can switch to
different states given the same current state and input symbol.  An NFSTA can,
therefore, have different runs on the same tree.  If at least one of the
possible runs is successful, the tree is accepted.

Similarly to the string language and classical tree cases, deterministic and
non-deterministic automata on \foo trees are equivalent: a language that is
recognised by an NFSTA is recognised by some DFSTA and vice versa.  The proof
of this and some other basic facts about automata will be given in
Section~\ref{sect:equiv} after a more formal definition of an FSTA run and
recognisable \foo tree languages is presented.

In order to formally define the run of a \foo tree automaton, we need to find
a way to associate current states with all the labels.	To do this, we shall
put the state in front of each label, separated by a special symbol, denoted
as a slash ($/$).  For example, $\langle\symb{abc}\rangle$ will be transformed
to $\langle q_0/\symb{abc}\rangle$, after which the automaton will consume
\symb{abc} step by step, changing the state symbol before the slash.  Thus,
intermediate trees will all belong to $T(Q \cup \{/\})$.  Note that $T(\A)
\subset T(Q \cup \{/\})$, because $\A \subseteq Q$.

\begin{definition}
\label{def:move_relation}
Let $\auto = (\A, Q, Q_f, q_0, \rules)$ be an FSTA; assume for convenience
that $Q$ does not contain slash: $/\not\in Q$.	Let also $a, b, c \in Q$;\: $r,
l \in (Q\cup\{\langle, \rangle, /\})\a$;\: and $s \in Q\a$.  The \emph{move
relation} \moverel between two trees from $T(Q\cup\{/\})$ is defined as
follows:
%\begin{enumerate}
%\renewcommand{\labelenumi}{(\alph{enumi})}
%\item\label{item:init} $l\langle s\rangle r	\:\moverel\:
%		l\langle q_0/s\rangle r$
%	\quad(``initial state assignment'');
%\item\label{item:horiz} $l\langle a/bs\rangle r	\:\moverel\:
%		l\langle c/s\rangle r$
%	\quad if $(a, b) \rightarrow c \in \rules$
%	\quad(``horizontal move'');
%\item\label{item:vert} $l\langle a/\rangle r	\:\moverel\:
%		lar$
%	\quad(``vertical move'').
%\end{enumerate}

\begin{center}
\mycounter{(\alph{mycnt})}
\renewcommand{\arraystretch}{1.2}
\begin{tabular}{lrclrr}
\cstep\label{step:init}  & $l\langle s\rangle r$	& \moverel &
		$l\langle q_0/s\rangle r$ &&
	\quad (initial state assignment) \\
\cstep\label{step:horiz} & $l\langle a/bs\rangle r$	& \moverel &
		$l\langle c/s\rangle r$ &
	if $(a, b) \rightarrow c \in \rules$ \quad &
	\quad (horizontal move) \\
\cstep\label{step:vert}  & $l\langle a/\rangle r$	& \moverel &
		$lar$ &&
	\quad (vertical move)
\end{tabular}
\end{center}
\Moverel{\ast} is the reflexive and transitive closure of \moverel.  A tree $t
\in T(\A)$ is \emph{accepted} by the automaton $\auto$ if there exists $q_f
\in Q_f$ such that $t \:\Moverel{\ast}\: \langle q_f/ \rangle$.  An empty tree
is therefore accepted if and only if $q_0 \in Q_f$.
\end{definition}

In the definition of the move relation, a tree from $T(Q\cup\{/\})$ contains
automaton's both input and state.  The role of slash is actually to mark those
labels to which step \ref{step:init} has already been applied.

As described above, an FSTA can make three different kinds of steps.  Step
\ref{step:init}---initial state assignment---applies once to each leaf label
(because $s$ cannot contain angle brackets or slashes).  Then, step
\ref{step:horiz}---horizontal move---transforms a label according to the rules
from $\rules$ by consuming one symbol and changing the state.  Finally, labels
which have been fully processed by step \ref{step:horiz} are cut off and their
final states are inserted into their parent labels in step
\ref{step:vert}---vertical move.  Eventually, intermediate nodes lose their
descendants and become leaves, making themselves available for step
\ref{step:init} and so on.  The process stops at the root (or when there is no
suitable rule in $\rules$).

\begin{example}
\label{exmp:automaton}
Let $\A = \{\symb{a}, \symb{b}\}$.  The following automaton $\auto = (\A, Q,
Q_f, q_0, \rules)$ accepts only trees whose labels (all of them) are composed
of the same letter, either \symb{a} or \symb{b}.
\begin{displaymath}
\begin{array}{lcl}
Q   & = & \{\symb{a}, \symb{b}, q_0\}	\\
Q_f & = & Q
\end{array}
\qquad
\rules = \left\{ 
\begin{array}{rcl}
(q_0, \symb{a})		& \rightarrow & \symb{a} \,, \\
(\symb{a}, \symb{a})	& \rightarrow & \symb{a} \,, \\
(q_0, \symb{b})		& \rightarrow & \symb{b} \,, \\
(\symb{b}, \symb{b})	& \rightarrow & \symb{b}
\end{array}
\right\}
\end{displaymath}
Consider the tree $\langle\langle\langle\symb{a}\rangle\rangle\symb{a}\langle
\symb{aa}\rangle\rangle$.  This tree is accepted by $\auto$, as illustrated
below by (slightly abbreviated) one of its possible runs:
\begin{eqnarray*}
&
\underset{\symb{a}}{\underset{\displaystyle|}{\varepsilon}}\raisebox{1.7ex}{$\diagup$}\raisebox{3.4ex}{\symb{a}}\raisebox{1.7ex}{$\diagdown$}\!\symb{aa}
\quad\Moverel{2}\quad
\underset{\displaystyle q_0/\symb{a}}{\underset{\displaystyle|}{\varepsilon}}\!\!\raisebox{1.7ex}{$\diagup$}\raisebox{3.4ex}{\symb{a}}\raisebox{1.7ex}{$\diagdown$}\!\!\!q_0/\symb{aa}
\quad\Moverel{2}\quad
\underset{\displaystyle\symb{a}/}{\underset{\displaystyle|}{\varepsilon}}\raisebox{1.7ex}{$\diagup$}\raisebox{3.4ex}{\symb{a}}\raisebox{1.7ex}{$\diagdown$}\!\symb{a}/\symb{a}
\quad\moverel\quad
\underset{\displaystyle\symb{a}/}{\underset{\displaystyle|}{\varepsilon}}\raisebox{1.7ex}{$\diagup$}\raisebox{3.4ex}{\symb{a}}\raisebox{1.7ex}{$\diagdown$}\!\symb{a}/
\quad\moverel
& \\ &
\parbox[][8.5ex]{0pt}{}
\raisebox{-1.7ex}{\symb{a}}\diagup\raisebox{1.7ex}{\symb{a}}\diagdown\raisebox{-1.7ex}{$\symb{a}/$}
\quad\moverel\quad
\raisebox{-1.7ex}{\symb{a}}\diagup\raisebox{1.7ex}{\symb{aa}}
\quad\moverel\quad
\raisebox{-1.7ex}{$q_0/\symb{a}$}\diagup\raisebox{1.7ex}{\symb{aa}}
\quad\moverel\quad
\raisebox{-1.7ex}{$\symb{a}/$}\diagup\raisebox{1.7ex}{\symb{aa}}
\quad\moverel
& \\ &
\parbox[][8.5ex]{0pt}{}
\symb{aaa}
\quad\moverel\quad
q_0/\symb{aaa}
\quad\moverel\quad
\symb{a}/\symb{aa}
\quad\moverel\quad
\symb{a}/\symb{a}
\quad\moverel\quad
\symb{a}/
&
\end{eqnarray*}

It is intuitively understandable that this automaton does not accept trees
containing mixed symbols: because of the way symbols are propagated, a mixed
tree would eventually ``resolve'' to the point where a leaf would contain
different symbols.  A move of the automaton on that leaf would require a rule
with $(\symb{a}, \symb{b})$ or $(\symb{b}, \symb{a})$ in its left-hand side,
but $\rules$ contains no such rule.  A complete proof of this statement is not
significant for the further discussion and is therefore omitted.
\end{example}

\subsection{Generalised automata}

Like with deterministic automata being a case of more general
non-deterministic automata, the latter can be reasonably generalised even
further.  In an NFSTA, non-determinism is only present in
step~\ref{step:horiz}---horizontal move.  Therefore, it seems natural to
extend non-determinism to the two other steps as well:
\begin{itemize}
\item Initial state assignment \ref{step:init} can be generalised by allowing
a set of possible initial states $Q_0$, rather than a single initial state
$q_0$.	
\item Vertical move \ref{step:vert} can be governed by a set of rules
$\gamma$, which (non-deterministically) map one state to another during the
move.
\end{itemize}
It happens that such generalisation
does not increase expressive power: all three kinds of \foo tree automata
recognise the same class of languages, as will be shown in
Section~\ref{sect:equiv}.  Because generalised automata are somehow cumbersome
to deal with, and because they are not as useful as their more restrictive
counterparts are, in our further study we shall be primarily dealing with
``simple'' FSTAs.  However, the concept of a generalised automaton will be
indispensable for proving the equivalence of automata and tree regular
grammars in Section~\ref{sect:rtgrammars}.

\begin{definition}
A Generalised Non-deterministic Finite \Foo Tree Automaton (GNFSTA) over an
alphabet $\A$ is a tuple
$\auto = (\A, Q, Q_f, Q_0, \rules, \gamma)$, where
\begin{itemize}
\item $Q \supseteq \A$ is a finite set, called \emph{set of states}, or
	\emph{state set};
\item $Q_f \subseteq Q$ is a set of final states;
\item $Q_0 \subseteq Q$ is a set of initial states;
\item $\rules$ is a set of ``horizontal'' transition rules of the form $(q_1,
q_2) \rightarrow q_3$, where $q_1,q_2,q_3 \in Q$;
\item $\gamma$ is a set of ``vertical'' transition rules of the form $q_1
\rightarrow q_2$, where $q_1, q_2 \in Q$.
\end{itemize}
It is often more convenient to treat $\gamma$ as a function $\gamma:\: Q
\rightarrow 2^Q$, so that $\gamma(q)$ denotes the set of states $q$ can be
transformed to during a vertical move.	This is the notation that will be used
in the rest of the paper.
\end{definition}

\begin{definition}
Let $\auto = (\A, Q, Q_f, Q_0, \rules, \gamma)$ be a GNFSTA; as usual, we
assume that $/\not\in Q$.  Let also $a, b, c \in Q$;\: $r, l \in
(Q\cup\{\langle, \rangle, /\})\a$;\: and $s \in Q\a$.  The \emph{move
relation} \moverel between two trees from $T(Q\cup\{/\})$ is defined as
follows:
\begin{center}
\mycounter{(\alph{mycnt})}
\renewcommand{\arraystretch}{1.2}
\begin{tabular}{lrcl@{\quad\quad}lr}
\cstep	& $l\langle s\rangle r$		& \moverel &
		$l\langle a/s\rangle r$ &
	if $a \in Q_0$ \quad &
	\quad (initial state assignment) \\
\cstep	& $l\langle a/bs\rangle r$	& \moverel &
		$l\langle c/s\rangle r$ &
	if $(a, b) \rightarrow c \in \rules$ \quad &
	\quad (horizontal move) \\
\cstep	& $l\langle a/\rangle r$	& \moverel &
		$lbr$ &
	if $b \in \gamma(a)$ \quad &
	\quad (vertical move)
\end{tabular}
\end{center}
As we can see from this definition, an empty tree is accepted if and only if
$Q_0 \cap Q_f \neq \emptyset$.
\end{definition}

\subsection{Locality}

As follows from the informal description of the operation of a bottom-up tree
automaton, the actions done in one branch of a tree are independent from the
actions performed in another branch.  Intuitively, the order in which
individual leaves are processed should not matter until their branches are
folded up to a common ancestor.

By definition, an automaton's run is sequential; there is no parallelism
allowed.  Thus, even a deterministic \foo tree automaton can produce different
runs on the same tree.	At each point there may be a choice of multiple labels
the next step can be applied to.  However, this fact does not really qualify
as non-determinism, because this choice does not affect the success of the run.

\begin{proposition}
\label{stmt:locality}
Consider a non-final tree in a successful GNFSTA run.  Then for any of its
leaves a step can be applied to that leaf that belongs to a (probably
different) successful run.
\end{proposition}

\begin{proof}
Let $\auto = (\A, Q, Q_f, Q_0, \rules, \gamma)$ be a GNFSTA that accepts tree
$t_0$.	Let $t \in T(Q \cup \{/\})$ be a non-final tree in the run of $\auto$
on $t_0$: $t_0 \Moverel{\ast} t \Moverel{n} \langle q_f/ \rangle$, where $n
\ge 1,\: q_f \in Q_f$.	Now, if $t$ consists of only one label, the statement
is trivial.  If $t$ contains more than one label, it can be represented as $t
= \langle x \langle w \rangle y \rangle$, where $w \in (Q \cup \{/\})\a$ is
one of its leaves.  We want to prove that for any such representation, $t
\moverel \langle xvy \rangle \Moverel{\ast} \langle q_f/ \rangle$ for some
$v$ such that:
$$
	v = \langle w' \rangle,\quad \text{where}\: w' \in (Q \cup \{/\})\a,
	\quad\quad \text{or} \quad\quad v = q,\quad \text{where}\: q \in Q.
$$
This is proven by induction on $n$.

\basis $n = 3$.  The smallest number of steps for a tree consisting of more
than one label is three:
$$
	\langle\langle q/ \rangle\rangle \moverel \langle q \rangle \moverel
	\langle q_0/q \rangle \moverel \langle q_f/ \rangle .
$$
In this case, $v = q$ and the statement is true.

\induction  Suppose the statement is true for $n$ steps; we need to prove it
for $n + 1$ steps.  Since $t \Moverel{n+1} \langle q_f/ \rangle$, there exists
$t'$ such that $t \moverel t' \Moverel{n} \langle q_f/ \rangle$.  Consider all
possible steps that can be applied to $t$.  By the move relation definition, 
the step from $t$ to $t'$ can be done on the label that is either $\langle w
\rangle$, or wholly contained within $x$ or $y$.  This gives us four
possibilities for this step (and thus for $t'$):
\begin{align*}
\langle x \langle w \rangle y \rangle & \moverel \langle x' \langle w
	\rangle y \rangle,
	& \langle x \langle w \rangle y \rangle & \moverel \langle x
	\langle w' \rangle y \rangle, \\
\langle x \langle w \rangle y \rangle & \moverel \langle x \langle w
	\rangle y' \rangle,
	& \langle x \langle w \rangle y \rangle & \moverel \langle xqy
	\rangle.
\end{align*}
If $t'$ is one of those listed in the right column, then the induction holds
immediately.  Otherwise, we assume that $t' = \langle x' \langle w \rangle y
\rangle$ (the other case is fully analogous).

Applying the inductive hypothesis to $t'$, we get $t' \moverel \langle x'vy
\rangle \Moverel{\ast} \langle q_f/ \rangle$.  The move from $t'$ to $\langle
x'vy \rangle$ is done via one of the three GNFSTA steps with $\:l = \langle
x'\:$ and $\:r = y\rangle\:$.  Let $l = \langle x$; then the same step will
apply to $t$:
$$
	t = \langle x \langle w \rangle y \rangle \moverel \langle xvy \rangle .
$$
Now, by analogy, consider the step from $t$ to $t'$.  During that step, $l$ is
a prefix of $\langle x$ and $r = z \langle w \rangle y$, where $z$ is an
arbitrary string from $(Q \cup \{\langle, \rangle, /\})\a$.  Let $r = zvy$;
then the same step will apply to the tree $\langle xvy \rangle$: $\langle xvy
\rangle \moverel \langle x'vy \rangle$.  Thus, we get:
$$
	t \moverel \langle xvy \rangle \moverel \langle x'vy \rangle
	\Moverel{\ast} \langle q_f/ \rangle,
$$
which proves the inductive statement.
\end{proof}

\subsection{Automata with ``pure'' states}
\label{sect:pure}

A salient difference between \foo tree automata and the traditional finite
state machines is that an FSTA's state set is a superset of the ``input''
alphabet $\A$.  In other words, in FSTAs all input symbols can serve as what
corresponds to states in traditional automata---for example, they can occur in
right hand sides of rules from $\rules$.

This came out of the fact that in \foo tree automata, ``traditional'' states
become input symbols during vertical moves.  Therefore, there is not much
conceptual difference between them; that's why we often call states
\emph{state symbols}.  An FSTA has to accept both sets as its input.  Then, to
make life easier and notation simpler, we allowed our FSTAs to use both sets
as states as well.  As a side effect, this has permitted in certain cases
simple and elegant implementations, such as the one from
Example~\ref{exmp:automaton}.

However, there is nothing special about using input symbols as states.	In
fact, any (GN)FSTA can be trivially rewritten so as not to use them.  This
rewrite preserves determinism of the automaton; i.e.\ whether it is
deterministic, non-deterministic, or generalised.

Indeed, let $\auto = (\A, Q, Q_f, Q_0, \rules, \gamma)$ be an FSTA.  Consider
a set of ``complementary'' symbols $\overline{\A}$ such that there is a unique
symbol $\bar{a} \in \overline{\A}$ for each $a \in \A$, and $\overline{\A}
\cap Q = \emptyset$.  Let the new state set $Q'$ be $Q \cup \overline{\A}$.
Let us also extend the complementation operator to $Q$ so that $\bar{q} = q$
if $q \in Q \smallsetminus \A$.  Then, for each rule from $\rules$ we add one
or two rules to (initially empty) $\rules'$ as follows:
$$
\rules' = \bigcup_{(q, r) \rightarrow p \in \rules}\left\{\begin{array}{ll}
	(\bar{q}, r)		\rightarrow \bar{p}, \\
	(\bar{q}, \bar{r})	\rightarrow \bar{p}
\end{array}\right\}
$$
(Note that when $r \not\in \A$, the set under the union sign is effectively a
single-element set.)  For a GNFSTA, we also build the new automaton's
$\gamma'$ as
$$
\begin{array}{ll}
    \gamma'(\bar{q}) = \overline{\gamma(q)} & \text{for all}\quad q \in Q; \\
    \gamma'(a) = \emptyset		    & \text{for all}\quad a \in \A.
\end{array}
$$
This procedure simply creates a duplicate set of input symbols, which can be
used as states, and then replaces input symbols with their duplicates in every
place where they were used as states.

Note the following properties of the automaton $\auto' = (\A, Q',
\overline{Q_f}, \overline{Q_0}, \rules', \gamma')$:
\begin{itemize}
\item no rule (horizontal or vertical) can produce a symbol from $\A$;
\item no rule takes a symbol from $\A$ as its state symbol;
\item neither the starting nor the final state sets contain symbols from $\A$.
\end{itemize}
Such automata will be called \emph{automata with pure states}.

Simple substitution by FSTA definition reveals that $L(\auto') = L(\auto)$.
Also, nothing in this procedure changes the deterministic properties of the
automaton.

\subsection{Equivalence of deterministic and non-deter\-mi\-nis\-tic automata}
\label{sect:equiv}

\begin{figure}[hbt]
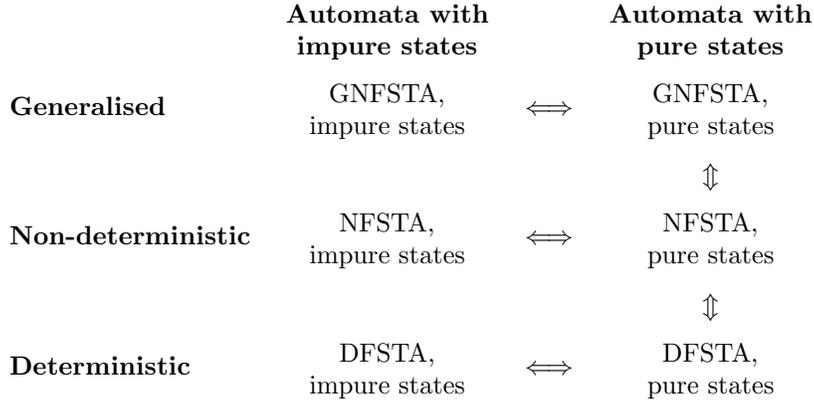

\begin{verse}
\renewcommand{\baselinestretch}{1}\normalsize
\renewcommand{\arraystretch}{1.5}
\begin{tabular}{lm{8em}cm{8em}}
		& \centering\bfseries Automata with\\impure states &
		& \centering\bfseries Automata with\\pure states\tabularnewline
\textbf{Generalised}	& \centering GNFSTA,\\impure states
	& $\Longleftrightarrow$
			& \centering GNFSTA,\\pure states	\tabularnewline
		&	&	& \centering $\Updownarrow$	\tabularnewline
\textbf{Non-deterministic} & \centering NFSTA,\\impure states
	& $\Longleftrightarrow$
			& \centering NFSTA,\\pure states	\tabularnewline
		&	&	& \centering $\Updownarrow$	\tabularnewline
\textbf{Deterministic}	& \centering DFSTA,\\impure states
	& $\Longleftrightarrow$
			& \centering DFSTA,\\pure states	\tabularnewline
\end{tabular}
\end{verse}
\caption{Equivalence of different kinds of \foo tree automata}
\label{fig:equiv}
\end{figure}

Now, we are ready to show that deterministic, non-deterministic, and
generalised automata recognise the same class of languages.
For this, it is sufficient to show that for any GNFSTA
there exists an equivalent (i.e., recognising the same language) DFSTA.
Moreover, as follows from the previous section, GNFSTAs can be assumed to be
automata with pure states.  The overview of the proof is presented in
Figure~\ref{fig:equiv}.

\begin{theorem}
\label{stmt:dg-equiv}
For any GNFSTA with pure states over $\A$ there exists an equivalent DFSTA over
$\A$.
\end{theorem}

This statement can be proven using the same technique as in the classical
(string) automata theory.  We only give an informal sketch of the proof here,
referring the reader to Appendix~\ref{app:equiv} for the complete proof.

Let $\auto$ be a GNFSTA $(\A, Q, Q_f, Q_0, \rules, \gamma)$ with pure states.
We construct DFSTA $\auto' = (\A', Q', Q'_f, q'_0, \rules')$, using the set of
all subsets of $Q$ as its state set.  All possible outcomes of rules from
$\rules$ with identical left hand side will then be lumped together into one
set; this set (which is a single state in the new automaton) will be assigned
to the corresponding deterministic rule in $\rules'$.

Firstly, however, one must remember that, according to the FSTA definition,
the state set must contains all the symbols from the alphabet.	We could
satisfy this by letting $Q' = 2^Q \cup \A$, but this would complicate the
proof unnecessarily.  Instead, we let the new alphabet be the set of all
single-element sets, containing symbols from the original alphabet: $\A' =
\big\{\{a\} \:|\: a \in \A\big\} = \bigcup_{a \in \A} \big\{\{a\}\big\}$.
Then we note that the natural bijection between $\A$ and $\A'$ implies a
one-to-one correspondence between $T(\A)$ and $T(\A')$, so the two automata
$\auto$ and $\auto'$ operate in fact on the same trees (with renamed alphabet
symbols).

Now, let $Q' = 2^Q$; \; $Q'_f = \{q' \in Q' \:|\: q' \cap Q_f \neq
\emptyset\}$; \; $q'_0 = Q_0$.	Because $\auto'$ does not have vertical rules,
their functionality has to be incorporated in horizontal rules.  To see how
this can be done, consider the following example that shows excerpts from a
GNFSTA run on a tree fragment:
\begin{center}
\renewcommand{\arraystretch}{1.8}
\begin{tabular}{m{7em}m{2em}m{7em}m{2em}m{8em}}
\centering $\underset{\displaystyle \symb{def}}{\underset{\displaystyle|}{\symb{ab}\,\symb{c}\;\;}}$ & \centering $\Moverel{\ast}$ &
\centering $\underset{\displaystyle q_1/}{\underset{\displaystyle|}{\symb{ab}\,\symb{c}\;\;}}$ & \centering $\Moverel{\ast}$ &
\centering ~\\ $q_{02}/\symb{ab}\,q_2\symb{c}$

\tabularnewline

\centering $t_0 = \langle\symb{ab}\langle\symb{def}\rangle\symb{c}\rangle$ & &
\centering $t_1 = \langle\symb{ab}\langle q_1/\rangle\symb{c}\rangle$	   & &
\centering $t_2 = \langle q_{02}/\symb{ab}\,q_2\symb{c}\rangle$
\end{tabular}
\end{center}
Here, $q_{02}$ is one of the starting states, and $q_2 \in \gamma(q_1)$.
Imagine an arbitrary leaf label in a GNFSTA run on some tree (looking at $t_2$
as an example).  Suppose it's ready for a horizontal move (i.e., contains /).
What symbols can it be composed of?
\begin{itemize}
\item The symbol on the left hand side of the / can be either an initial
state, or a result of some horizontal rule.
\item The symbols on right hand side of the / can be either original input
symbols from $\A$ (such as \symb{abc}), or results of some vertical rule (such
as $q_2$, which was produced by the rule $q_1 \rightarrow q_2$ from $\gamma$).
\end{itemize}
If vertical rules were not allowed, then $q_1$ would have squeezed into $t_2$
in the place of $q_2$.	Thus, new horizontal rules from $\rules'$ have to
apply $\gamma$ to all right hand side symbols that result from vertical
moves, prior to looking up an appropriate horizontal rule from $\rules$.
Roughly speaking, $\rules'(p, q) = \rules(p, \gamma(q))$, if $q$ came via a
vertical move from a subordinate node; and $\rules'(p, q) = \rules(p, q)$, if
$q$ is an input symbol that originally belonged to the node being processed.

Fortunately, distinguishing between these two cases is very easy, because our
GNFSTA is an automaton with pure states.  Indeed, in such an automaton
$\gamma$ is not allowed to produce symbols from $\A$, which is where all the
original input symbols come from.  Thus, if a symbol in a label belongs to
$\A$, it must be an original one; otherwise, the symbol has to be a result of
a vertical rule.  Appendix~\ref{app:equiv} gives a full proof why automaton
$\auto'$ built as described above is indeed equivalent to $\auto$.

\subsection{Finite \foo tree automata and classical automata}

Due to their historical background, FSTAs share much in common with classical
string and tree automata.  In a sense, FSTAs can be thought of as a
generalisation of classical automata: an FSTA can be applied
wherever a string or tree automaton is used.  The aim of this section is to
discuss the relationships between different kinds of automata.

\subsubsection{String automata}

\emph{Finite automata} (FA), also called \emph{finite state machines} (FSM),
are widely used in automata and formal language theories.  In this paper, we
often call them ``finite string automata'' to explicitly distinguish them from
tree automata.	A finite automaton is a tuple $(\A, Q, Q_f, q_0, \delta)$,
where $\A$ is an alphabet, $Q$ is a finite state set, $Q_f \subseteq Q$ is a
set of final states, $q_0 \in Q$ is the initial state, and $\delta$ is a map
from $Q \times \A$ to $2^Q$.  A finite automaton works on strings from $\A\a$,
starting in the state $q_0$ and applying rules from $\delta$ to its current
state and the next symbol from the input string to determine its next state.
The language recognised by an automaton is the set of strings that it accepts.

The move of an FA is, therefore, defined exactly like the combination of steps
\ref{step:init} and \ref{step:horiz} of the FSTA move (see
Definition~\ref{def:move_relation}).  This leads us to the following
statement.

\begin{proposition}
If $L$ is the language recognised by a finite automaton $(\A, Q, Q_f, q_0,
\delta)$, then $\langle L \rangle$ is recognised by the FSTA $(\A, Q \cup \A,
Q_f, q_0, \delta)$, where $\delta$ is naturally extended so that $\delta(q_1,
q_2) = \emptyset$ for all $q_1, q_2 \in Q$.
\end{proposition}

As we remember, $\langle L \rangle = \{ t \in T(\A) \:|\: t = \langle s
\rangle \quad \text{for some } s \in L \}$.  Indeed, for single-label trees
this statement follows immediately from the automata definitions.  Trees
containing more than one label are not accepted by this FSTA for the following
reason.
Suppose $t \in T(\A)$ has more than one label.	Then there exist a pair of
labels $l_1$ and $l_2$ such that $l_1$ contains $l_2$: $l_1 = \langle x l_2 y
\rangle$.  When $l_2$ has been fully processed, it becomes $\langle q_2/
\rangle$, where $q_2 \in Q$ by definition of the FA (because $q_0 \in Q$ and
Im\,$\delta \subseteq 2^Q$).  A vertical move then injects $q_2$ into $l_1$:
$\langle x q_2 y$.  The automaton will stop at $q_2$ because there is no
suitable rule in $\delta$.

A number of finite automata can be combined together to form an FSTA that
accepts different string languages in labels depending on their position in
the tree, where position is determined by the topology of the tree and by the
information in other labels.  This will be discussed in more details in
Section~\ref{sect:rtgrammars}.

\subsubsection{Tree automata}

In this paper, these are usually referred to as ``classical tree automata''
(CTA).	The following introduction is largely borrowed from~\cite{tata}.

Trees in the classical tree automata theory are called \emph{terms} and are
composed of \emph{ranked symbols}.  A \emph{ranked alphabet} is a finite set
of symbols, in which each symbol is associated with a whole non-negative
number, called \emph{arity}.  The arity of a symbol $f$ is denoted as
$\mathrm{Arity}(f)$.  Symbols of arity $0, 1, 2, \ldots, p$ are called
\emph{constants}, \emph{unary}, \emph{binary}, \ldots, \emph{$p$-ary symbols},
respectively.

The set $\hat{T}(\mathcal{F})$ of terms over the ranked alphabet $\mathcal{F}$
is the smallest set defined by:
\begin{itemize}
\item $f \in \hat{T}(\mathcal{F})$\quad for any constant $f \in \mathcal{F}$;
and
\item if $f \in \mathcal{F}$, $p = \mathrm{Arity}(f) \ge 1$, and $t_1, \ldots,
t_p \in \hat{T}(\mathcal{F})$, then $f(t_1, \ldots, t_p) \in
\hat{T}(\mathcal{F})$.
\end{itemize}
For example, consider a ranked alphabet $\mathcal{F}$ consisting of a binary
symbol $+$ and two constants $1$ and $2$: $\mathcal{F} = \{ +(,),\, 1,\, 2
\}$.  A term $+(1,+(1,2))$ represents the following tree:
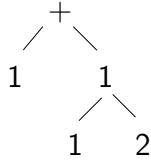
\begin{center}
\input{classical-tree.ltp}
\end{center}
A term $\hat{t} \in \hat{T}(\mathcal{F})$ may be viewed as a finite ordered
labelled tree, the leaves of which are labelled with constants and the
internal nodes are labelled with symbols of positive arity.  The number of
children a node has must be equal to the arity of the node's symbol.

Consequently, a term from $\hat{T}(\mathcal{F})$ can be represented as a \foo
tree over the un-ranked alphabet $\A = \mathcal{F}$.  This is illustrated by
the following recursively defined injective map $\tau : \hat{T}(\mathcal{F})
\rightarrow T(\A)$:
\begin{itemize}
\item $\tau(a) = a$ \quad (for a constant symbol $a$);
\item $\tau(f(t_1, \ldots, t_p)) = \langle f \tau(t_1) \cdots \tau(t_p)
	\rangle$.
\end{itemize}
For instance, the term $+(1,+(1,2))$ pictured above would map to
$\langle\symb{+}\langle\symb{1}\rangle\langle\symb{+}\langle\symb{1}\rangle
\langle\symb{2}\rangle\rangle\rangle$.

A non-deterministic classical finite tree automaton (NCFTA) over $\mathcal{F}$
is a tuple $\auto = (Q, \mathcal{F}, Q_f, \rules)$.  $Q$ is an alphabet,
consisting of constant symbols called states; $Q_f \subseteq Q$ is a set of
final states.  $\rules$ is a set of rules of the following type: $f(q_1,
\ldots, q_n) \rightarrow q$, where $n = \mathrm{Arity}(f)$, $q, q_1, \ldots,
q_n \in Q$.

Like our \foo tree automata, a classical bottom-up tree automaton also starts
at the leaves and moves upwards, associating a state with each subterm.  If
the direct subterms $u_1, \ldots, u_n$ of term $t = f(u_1, \ldots, u_n)$ are
assigned states $q_1, \ldots, q_n$ respectively, then the term $t$ will be
assigned some state $q$ given that $f(q_1, \ldots, q_n) \rightarrow q \in
\rules$.  The move relation $\Moverel[\hat{\auto}]{}$ is thus defined in
$\hat{T}(\mathcal{F} \cup Q)$; its full formal definition is outside the scope
of this paper and can be found in \cite{tata}.

Tree languages recognised by classical automata happen to be recognisable by
\foo tree automata, as demonstrated by the following proposition.

\begin{proposition}
Let $\hat{\auto}$ be an NCFTA $(\hat{Q}, \mathcal{F}, \hat{Q}_f,
\hat{\rules})$, recognising language $\hat{L} \subseteq \hat{T}(\mathcal{F})$,
and let $\A$ be the (non-ranked) set of symbols of the alphabet $\mathcal{F}$.
Then there exists an NFSTA $\auto = (\A, Q, Q_f, q_0, \rules)$ that recognises
the language $L = \tau(\hat{L})$.
\end{proposition}

\begin{proof}
The automaton $\auto$ is constructed step-by-step, by taking transition rules
from $\hat{\rules}$ and populating $Q$ and $\rules$ as described below.

Let $\rules_0 = \emptyset$,\, $Q_f = \hat{Q}_f$,\, and $Q_0 = \hat{Q} \cup
\{q_0\}$,\, where $q_0$ is a new state symbol ($q_0 \not\in \hat{Q}$).	Let us
enumerate the rules in $\hat{\rules}$, indexing them from $1$ to $n$.

For each $i$ from $1$ to $n$ we do the following:
assuming that the $i$-th rule in $\hat{\rules}$ is $f_i(\hat{q}_{i1}, \ldots,
\hat{q}_{ip_i}) \rightarrow \hat{q}_i, \; p_i \ge 0$,
let
\begin{displaymath}
Q_i = Q_{i-1} \cup \{ q_{i1}, \ldots, q_{ip_i} \}, \quad
\rules_i = \rules_{i-1} \cup \left\{
	\begin{array}{rcl}
		(q_0, f_i)			 & \rightarrow & q_{i1},   \\
		(q_{i1}, \hat{q}_{i1})		 & \rightarrow & q_{i2},   \\
						 & \vdots      &	   \\
		(q_{i,p_{i-1}}, \hat{q}_{i,p_{i-1}}) &\rightarrow& q_{ip_i}, \\
		(q_{ip_i}, \hat{q}_{ip_i})	 & \rightarrow & \hat{q}_i \\
	\end{array}
\right\},
\end{displaymath}
where $q_{ij}$ ($1 \le i \le n,\, 1 \le j \le p_i$) are new unique states
($q_{ij} \not\in Q_{i-1}$).

Finally, let $Q = Q_n$, $\rules = \rules_n$.  The language, recognised by the
resulting NFSTA $\auto = (\A, Q, Q_f, q_0, \rules)$, is exactly (up to the
mapping $\tau$) the language recognised by the original NCFTA, which can
be proven by applying the relevant definitions.
\end{proof}

%\begin{proposition}
%Let $\auto$ be an NFSTA $(\A, Q, Q_f, q_0, \rules)$, accepting language $L
%\subseteq T(\A)$, and $\mathcal{F}$ be a ranked alphabet consisting of symbols
%from $\A$.  Then there exists an NCFTA $\hat{\auto} = (\hat{Q}, \mathcal{F},
%\hat{Q}_f, \hat{\rules})$ that accepts language $\hat{L}$ such that $L \cap
%\tau(\hat{T}(\mathcal{F})) = \tau(\hat{L})$.
%\end{proposition}
%
%\begin{proof}
%I don't know whether I should waste my time proving this insignificant result.
%It feels quite obvious---everything's regular and stuff---but the actual proof
%may be somehow tedious.
%\end{proof}

\subsubsection{Vertical move and classical string automata}

As already discussed, the horizontal move of a \foo tree automaton is an exact
copy of the step of a traditional string automaton.  Let us now investigate the
vertical move from this point of view.

Consider string \symb{abc} and tree $\langle\langle\langle\langle
\symb{a}\rangle\symb{b}\rangle\symb{c}\rangle\rangle$.	The tree is arranged
so that one vertical move is required for an FSTA to consume each symbol,
bringing the analogy between an FSTA's vertical movement and a step of a
string automaton.  Note the seemingly excessive extra pair of angle brackets
in the middle; they actually simplify things, as will be shown below.
Despite the fact that the actual moves are quite different---for example, an
FSTA has to start with its initial state at each symbol, while a conventional
FA only starts with $q_0$ at the beginning of a string---any FA can be
implemented as an FSTA working with ``vertical'' tree representations of
strings.

Let $\A$ be an alphabet.  We shall call a tree $t \in T(\A)$ a \emph{vertical
tree representation} of a string $s \in \A\a$, if
\begin{displaymath}
s = a_1\cdots a_n \quad (a_1,\ldots,a_n \in \A,\: n\ge 0)
\mbox{\quad and \quad}
t = \underbrace{\langle\langle\cdots\langle}_{n+1} \rangle a_1\rangle
a_2\rangle \cdots a_n\rangle\;.
\end{displaymath}
The vertical tree representation can be obtained as the image of the function
$\omega : \A\a \rightarrow T(\A)$, recursively defined as
\begin{itemize}
\item $\omega(\varepsilon) = \langle\rangle$ \quad and
\item $\omega(sa) = \langle\omega(s)\rangle \cdot \langle a\rangle$ \quad
(where $a \in \A$, $s \in \A\a$).
\end{itemize}

%A non-deterministic finite automaton (NFA) over \A is a tuple $\tilde{\auto} =
%(\tilde{Q}, \A, \tilde{Q}_f, q_0, \tilde{\rules})$, where $\tilde{Q}$ is a
%finite set of states, $\tilde{Q}_f \subseteq \tilde{Q}$ is the set of final
%states, $q_0 \in \tilde{Q}$ is the initial state, and $\tilde{\rules}$ is a
%set of transition rules of the form:
%$$
%	(q, a) \rightarrow p \quad (q, p \in \tilde{Q};\; a \in \A)\:.
%$$
%We use the tilde symbol ($\tilde{ }$) to mark elements pertaining to finite
%string automata (as opposed to tree automata), to help us distinguish them in
%future.  A run of an FA on string $s$ is identical to a run of an FSTA on
%string $\langle s\rangle$; the move relation $\Moverel[\tilde{\auto}]{}$ is
%defined as
%$$
%	(q, aw) \Moverel[\tilde{\auto}]{} (p, s), \quad\mbox{where} q, p \in
%	\tilde{Q};\: a \in \A;\: s \in \A\a\,.
%$$

\begin{proposition}
For any FA over $\A$ there exists an FSTA over $\A$ that recognises the
vertical tree representation of the FA's language.
\end{proposition}

\begin{proof}[Sketch of proof]
Let $\fauto = (\A, Q, Q_f, q_0, \delta)$ be a finite string
automaton, recognising language $L(\fauto) \subseteq \A\a$.  We want to
construct a finite \foo tree automaton $\auto$ that recognises the language
$\omega(L(\fauto)) \subset T(\A)$.

To build $\auto$, we need an additional set of complementary states
$\overline{Q}$ such that there is a unique state $\bar{q} \in \overline{Q}$
for each $q \in Q$, where $\bar{q} \not\in Q,\, \bar{q} \not\in \A$.  We also
need a new unique initial state $q'_0$.  Each step of $\fauto$ will map to
four steps of $\auto$: horizontal, vertical, initial state assignment, and
horizontal again.  The first horizontal move will each time produce a
complementary state; vertical move will deliver it one level up; and the
following two steps will convert this complementary state to its counterpart,
preparing for the next horizontal move.  This is illustrated below by a run of
some FA on the string \symb{ab} and a corresponding run of an FSTA on
$\omega(\symb{ab}) = \langle\langle\langle\rangle \symb{a} \rangle \symb{b}
\rangle$:

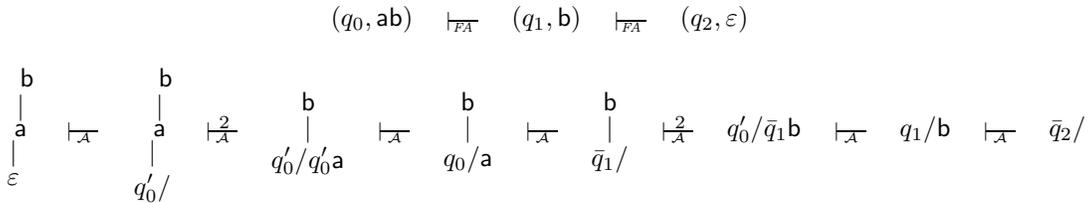
\newcommand{\vtreee}[3]{%
    \underset{\displaystyle#1}{\underset{\displaystyle|}{\;\;\overset{%
    \;\;\displaystyle#3}{\overset{\displaystyle|}{\displaystyle#2}}}}%
}
\newcommand{\vtree}[2]{%
    \underset{\displaystyle#1}{\overset{%
    \displaystyle#2}{\displaystyle|}}%
}

\begin{eqnarray*}
&
(q_0, \symb{ab}) \quad\Moverel[\fauto]{}\quad (q_1, \symb{b})
\quad\Moverel[\fauto]{}\quad (q_2, \varepsilon)
& \\[2ex] &
\vtreee{\varepsilon}{\symb{a}}{\symb{b}}
\quad\Moverel{}\quad
\vtreee{q'_0/}{\symb{a}}{\symb{b}}
\quad\Moverel{2}\quad
\vtree{q'_0/q'_0\symb{a}}{\symb{b}}
\quad\Moverel{}\quad
\vtree{q_0/\symb{a}}{\symb{b}}
\quad\Moverel{}\quad
\vtree{\bar{q}_1/}{\symb{b}}
\quad\Moverel{2}\quad
q'_0/\bar{q}_1\symb{b}
\quad\Moverel{}\quad
q_1/\symb{b}
\quad\Moverel{}\quad
\bar{q}_2/
&
\end{eqnarray*}

Let $\auto = (\A, Q', Q'_f, q'_0, \rules)$, where
\begin{align*}
Q'   & = Q \cup \overline{Q} \cup \{q'_0\} \cup \A, &
Q'_f & = \left\{\begin{array}{lll}
	\overline{Q_f}			& \text{if}	& q_0 \not\in Q_f, \\
	\overline{Q_f} \cup \{q'_0\}	& \text{if}	& q_0 \in Q_f,
\end{array}\right.
\end{align*}
and the rule set $\rules$ is built as follows:
\begin{itemize}
\item for each rule $(q, a) \rightarrow p$ from $\delta$ we add
	$(q, a) \rightarrow \bar{p}$ to $\rules$;
\item for each state $q \in Q$ we add
	$(q'_0, \bar{q}) \rightarrow q$ to $\rules$;
\item $\rules$ also contains the rule $(q'_0, q'_0) \rightarrow q_0$:
\end{itemize}
$$
	\rules = \bigcup_{(q,a)\rightarrow p \,\in\, \delta}
		\{ (q, a) \rightarrow \bar{p} \} \quad\cup\quad
		\bigcup_{q \in Q} \{ (q'_0, \bar{q}) \rightarrow q \}
		\quad\cup\quad \{ (q'_0, q'_0) \rightarrow q_0 \}
$$
Let $L(\auto)$ be the language recognised by $\auto$.  To prove the equivalence
of the two automata, we must show that
\begin{enumerate}
\renewcommand{\theenumi}{(\roman{enumi})}
\renewcommand{\labelenumi}{\theenumi}
\item\label{item:fatovert}
	$s \in L(\fauto)$ implies that $\omega(s) \in L(\auto)$;\; and
\item\label{item:verttofa}
	for any $t \in L(\auto)$ there exists $s \in L(\fauto)$ such that
	$\omega(s) = t$.
\end{enumerate}

Statement~\ref{item:fatovert} can be proven by considering the run of $\fauto$
on $s$ and building the corresponding run on $\omega(s)$, replacing each
original step with four new as illustrated above.  The result is then shown to
be a valid successful run of $\auto$.

To prove statement~\ref{item:verttofa}, consider the sequence of tree sets
$T_n$, composed of trees that produce a successful run in $n$ steps:
$T_n = \{ t \in T(Q' \cup \{/\}) \:|\: t \Moverel{n} \langle q'_f\rangle,\,
q'_f \in Q'_f \}$.  Induction on $n$ with 4 step increments shows that all
trees from $L(\auto)$ are contained in $T_{4k+1}$ ($k = 0, 1, \dots$) and that
each tree from $L(\auto) \cap T_{4k+1}$ can be represented as $\omega(s)$,
where $s \in L(\fauto)$.
\end{proof}

\subsubsection{Boolean closure}

According to the formal language theory, the class of recognisable (regular)
string languages is closed under union, under intersection, and under
complementation~\cite{hopul:automata-theory}.  The same applies to the
classical tree languages~\cite{tata}.  Quite naturally, \foo tree languages
are no exception and also exhibit the same Boolean closure properties.	The
proof from classical literature can be easily transferred to our case.	Thus,
only the main idea of the proof is presented here.

To prove that languages $L_1 \cup L_2$ and $L_1 \cap L_2$ are recognisable, we
take the FSTAs that recognise $L_1$ and $L_2$ and unite their state and rule
sets.  The final state sets for the new automata are taken as the union and
intersection, respectively, of the original final state sets.

To prove that and $T(\A) \smallsetminus L$ is recognisable, we consider a
complete FSTA that recognises $L$.  Any FSTA can be made complete, as shown in
Corollary~\ref{stmt:complete} in Appendix~\ref{app:equiv}.  Complementing its
final state set produces the automaton that recognises $T(\A) \smallsetminus
L$.

%TODO: discussion of the horizontal and vertical runs?
%In case of reduced trees we can mix symbols and states...

\subsection{Discussion}

We have compared regular \foo trees with regular strings and classical trees.
The most important conclusion is that the former can ``grow'' in two
directions: horizontally and vertically, combining the properties of the
latter two.  Classical trees, for instance, can regularly grow downwards, but
the degree of each node is fixed and depends on the node's symbol (label).  In
\foo trees, a node can have a regular set of child trees.

The noted similarities between traditional regular sets and horizontal and
vertical arrangements of nodes in regular \foo trees hint that usual regular
properties are likely to hold.	Pumping, for example, applies to trees growth
both in width and in depth.

\section{Regular languages, grammars, and expressions}
\label{sect:grammars}

In this section, we discuss an alternative approach to the definition of
recognisable tree languages, based on the concept of grammars.	Again, this
goes in parallel with, and derives from, the classical theories.

\subsection{Regular tree grammars}
\label{sect:rtgrammars}

In contrast with an automaton, which is an accepting device, a grammar is a
generating device.  A grammar defines set of rules, which generate objects
(strings or trees) from a pre-defined starting point.  The language defined by
a grammar is the set of all objects that can be generated using the rules of
the grammar.

A \foo tree grammar is very similar to an extended context-free grammar (an
extended CFG is a CFG which allows regular expressions, rather than simple
sequences, in the right hand sides of its rules; ECFGs have the same
expressive power as normal CFGs).  Before proceeding with the tree grammar
definition, we need to briefly introduce ``string-regular'' expressions, which
we shall be using extensively.

A \emph{regular expression} (RE) \cite{hopul:automata-theory} is a mechanism
of formal language theory that describes regular languages.  A regular
expression over $\A$ defines a (regular) subset of $\A\a$, using symbols from
$\A$ and three \emph{regular operators}.  The set $\RE(\A)$ of regular
expressions over $\A$ is defined as follows:
\begin{itemize}
\item $\varepsilon \in \RE(\A)$;
\item $\A \subset \RE(\A)$;
\item if $\,r_1, r_2 \in \RE(\A)$,\, then $\,r_1r_2 \in \RE(\A)$ \quad
	(concatenation);
\item if $\,r_1, r_2 \in \RE(\A)$,\, then $\,r_1 | r_2 \in \RE(\A)$ \quad
	(union);
\item if $\,r \in \RE(\A)$,\, then $\,r\a \in \RE(\A)$ \quad
	(iteration, or Kleene star).
\end{itemize}
The language defined by a RE $r$ is denoted as \reglang{r}.

\begin{definition}
A regular \foo tree grammar (RSTG) is a tuple $(\A, N, S, R)$, where:
\begin{itemize}
\item $\A$ is a finite alphabet,
\item $N$ is a finite set of \emph{non-terminal} symbols ($N\cap\A =
	\emptyset$),
\item $S \in N$ is an \emph{axiom}, or \emph{starting non-terminal}, and
\item $R$ is a set of \emph{production rules} of the form $n \rightarrow
	\langle r\rangle$, where $n \in N$, $\,r \in \RE(\A \cup N)$.
\end{itemize}
\end{definition}
The only difference between RSTGs and extended context-free grammars is the
pair of angle brackets in the right hand sides of production rules.  They
are there to indicate that every application of a production rule inserts a
pair of angle brackets into the generated string.

The \emph{derivation relation} $\derivrel$, associated to a regular tree
grammar $G = (\A, N, S, R)$, is defined on pairs of strings from $(\A \cup N
\cup \{\langle,\rangle\})\a$:
$$
	u \derivrel v \;\;\Longleftrightarrow\;\; \left\{\begin{array}{lll}
		u = lXr, & X \rightarrow e \in R & \quad\text{and} \\
		v = l\langle x\rangle r, & x \in \reglang{e},
	\end{array}\right.
$$
where $u, v, l, r \in (\A \cup N \cup \{\langle,\rangle\})\a$,\: $X \in N$,\:
$x \in (\A \cup N)\a$,\: $e \in \RE(\A \cup N)$.

\begin{theorem}
%Regular tree grammars on \foo trees describe the same class of languages as
%finite tree automata.
A \foo tree language is recognisable (by a finite \foo tree automaton) if and
only if it is generated by a regular tree grammar.
\end{theorem}

\begin{proof}
\textbf{Grammar $\rightarrow$ automaton:}
Given some regular tree grammar $G = (\A, N, S, R)$, we show how to build a
generalised finite tree automaton with pure states $\auto = (\A, Q, Q_f, Q_0,
\rules, \gamma)$ which recognises $L(G)$---the language generated by $G$.

Let $k = |R|$ be the number of production rules in $R$.  Each rule $r_i \in R$
has the form $n_i \rightarrow \langle e_i \rangle$, where $n_i$ is a
non-terminal from $N$ and $e_i$ is a regular expression over $\A \cup N$.
According to the automata theory, for each regular expression there exists a
finite (string) automaton that recognises the same language.  Let
$\mathrm{FA}_i$ be a finite automaton equivalent to $e_i$.
Consider $\mathrm{FA}_i = (\A \cup N, Q_i, q_{0i}, Q_{fi}, \delta_i)$ for $i =
1, \dots, k$, where:
\begin{itemize}
\item $Q_i$ is a finite set of states, disjoint with $\A \cup N$;
\item $q_{0i}$ is the initial state, $q_{0i} \in Q_i$;
\item $Q_{fi}$ is the set of final states, $Q_{fi} \subseteq Q_i$;
\item $\delta_i$ is a map from $Q_i \times (\A \cup N)$ to $2^Q$;
or---equivalently---a set of rules of the form $(q, a) \rightarrow p$, where
$q, p \in Q,\: a \in \A \cup N$.
\end{itemize}
Since all these automata are independent (apart from having a common alphabet
$\A \cup N$), we can assume that their state sets $Q_1, \dots, Q_k$ are
pairwise disjoint.

We now build our GNFSTA $\auto = (\A, Q, Q_f, Q_0, \rules, \gamma)$ as follows:
\begin{eqnarray*}
       Q & = & \A \cup N \cup Q_1 \cup \dots \cup Q_k		\\
     Q_0 & = & \{ q_{01}, \dots, q_{0k} \}			\\
  \rules & = & \delta_1 \cup \dots \cup \delta_k		\\
  \gamma & = & \bigcup_{i = 1, \dots, k} {\textstyle\bigcup\limits_{q \in Q_{fi}}}
		\{ q \rightarrow n_i \}				\\
     Q_f & = & \gamma^{-1}(\{S\}) = \bigcup_{i:\: S \rightarrow
		\langle e_i \rangle \in R} Q_{fi}
\end{eqnarray*}
To prove the equivalence of $G$ and $\auto$, we show for any tree $t \in T(\A
\cup N)$ that it is generated by $G$ if and only if it is accepted by $\auto$.
The induction is done on the number of labels in $t$.

\basis $n = 1$.  Then $t = \langle s \rangle$, where $s \in (\A \cup N)\a$.

\there If $t$ is generated by $G$, then there is a rule $S \rightarrow \langle
e_i \rangle$ in $R$ such that $s \in \reglang{e_i}$.  By definition of
$\auto$, $s$ is accepted by $\mathrm{FA}_i = (\A \cup N, Q_i, q_{0i}, Q_{fi},
\delta_i)$, where $Q_i \subset Q,\, q_{0i} \in Q_0,\, Q_{fi} \subseteq Q_f,\,
\delta_i \subseteq \rules$.  Therefore, $\langle s \rangle$ is accepted by
$\auto$ as follows:
$$
	\langle s \rangle \moverel \langle q_{0i}/s \rangle \Moverel{\ast}
	\langle q_f/ \rangle
$$
for some $q_f \in Q_{fi} \subseteq Q_f$.

\back If $\langle s \rangle$ is accepted by $\auto$, then there is a run
$$
	\langle s \rangle \moverel \langle q_0/s \rangle \moverel \langle
	q_1/s_1 \rangle \moverel \cdots \moverel \langle q_f/ \rangle,
$$
where $q_0 \in Q_0;\: s = a_1s_1,\, s_1 = a_2s_2, \ldots$; $(q_{j-1}, a_j)
\rightarrow q_j \,\in\, \rules$; and $q_f \in Q_f$.  By definition of $Q_0$,
$q_0 = q_{0i} \in Q_i$ for some $i$.  Because $Q_1, \dots, Q_k$ are pairwise
disjoint (and by definition of $\rules$), the rule $(q_0, a_1) \rightarrow
q_1$ belongs to $\delta_i$, which implies that $q_1 \in Q_i$.  The same
reasoning can then be applied to $q_1, q_2, \ldots$, showing that all the
horizontal rules applied belong to the same $\delta_i$ and $q_f \in Q_{fi}$.
Thus, $s$ is accepted by the finite string automaton $\mathrm{FA}_i$.  (If $s$
is empty, then $q_0 \in Q_f$, therefore $q_0 \in Q_{fi}$ for some $i$, so $s$
is accepted by $\mathrm{FA}_i$.)

Notice that $q_f$ belongs to both $Q_{fi}$ and $Q_f$, which by definition of
$Q_f$ implies that the rule $S \rightarrow \langle e_i \rangle$ belongs to
$R$, where $e_i$ is the regular expression equivalent to $\mathrm{FA}_i$.
Thus, the grammar $G$ generates $\langle \reglang{e_i} \rangle$, and in
particular, $\langle s \rangle$.

\induction  By inductive hypothesis, we assume that any tree $t' \in T(\A \cup
N)$ that has $m$ labels ($m \ge 1$) is generated by $G$ if and only if it is
accepted by $\auto$.  We want to show that the same holds true for any tree
$t$ with $m + 1$ labels.  Let us select a leaf label in $t$: $t = l \langle s
\rangle r$, where $s \in (\A \cup N)\a$.

\there If $t$ is produced by $G$, then there is a tree $t' = lnr$ also
produced by $G$ such that $n \in N$ and there is a rule $r_i:\: n \rightarrow
\langle e \rangle$ in $R$ such that $s \in \reglang{e}$.  Thus, by definition
of $\auto$,
$$
	l \langle s \rangle r \;\moverel\; l \langle q_{0i}/s \rangle r
	\;\Moverel{\ast}\; l \langle q_f/ \rangle r
$$
for some $q_f \in Q_{fi}$.  Then we notice that $\gamma$ contains the rule
$q_f \rightarrow n$, which implies $l \langle q_f/ \rangle r \moverel lnr$.
By induction, $lnr = t'$ is accepted by $\auto$, therefore $t$ is accepted as
well.

\back If $t = l \langle s \rangle r$ is accepted by $\auto$, then by
Proposition~\ref{stmt:locality} (locality) there is a successful run of
$\auto$ on $t$:
$$
	l \langle s \rangle r \;\moverel\; l \langle q_0/s \rangle r
	\;\moverel\; l \langle q_1/s_1 \rangle r \;\moverel\; \cdots
	\;\moverel\; l \langle q_z/ \rangle r \;\moverel\; lq'r
	\;\Moverel{\ast}\; \langle q_f/ \rangle,
$$
where $q_0 \in Q_0;\: s = a_1s_1,\, s_1 = a_2s_2, \ldots$; $(q_{j-1}, a_j)
\rightarrow q_j \,\in\, \rules$; $q_z \rightarrow q' \,\in\, \gamma$; and
$q_f \in Q_f$.  By analogy with the inductive basis, we can see that $q_0 =
q_{0i}$ for some $i$; all rules $(q_{j-1}, a_j) \rightarrow q_j$ belong to
$\delta_i$ for the same $i$; $q_z \in Q_{fi}$; and $q' = n_i$ (the
non-terminal in the left hand side of the $i$-th rule in $R$).  Thus, $s$ is
accepted by $\mathrm{FA}_i$, equivalent to the regular expression $e_i$ in the
rule $n_i \rightarrow \langle e_i \rangle \,\in\, R$.

Consider the tree $ln_ir$.  It is accepted by $\auto$ (because $ln_ir = lq'r$
and $lq'r$ is accepted), therefore by the inductive hypothesis $ln_ir$ is
generated by $G$.  Then, all trees $l \langle \reglang{e_i} \rangle r$ are
also generated by $G$; and this includes $t = l \langle s \rangle r$, because
$s \in \reglang{e_i}$.

\medskip\noindent\textbf{Automaton $\rightarrow$ grammar:}
Given a finite tree automaton with pure states $\auto = (\A, Q, Q_f, q_0,
\rules)$, we want to build a regular tree grammar $G = (\A, N, S, R)$ that
recognises $L(\auto)$.

Let $N = Q \smallsetminus \A \cup \{S\}$, where $S$ is a symbol not in $Q$ or
$\A$.  For each $n_i \in N \smallsetminus \{S\}$, let us build an automaton
$\mathrm{FA}_i = (Q, Q \smallsetminus \A, \{n_i\}, q_0, \rules)$, whose input
symbols are all the states of $\auto$, and whose states are the ``pure
states'' $(Q \smallsetminus \A)$ of $\auto$.  The transition rules and the
initial state for each $\mathrm{FA}_i$ are taken directly from $\auto$.
%(this can be done because $\auto$ is an automaton with pure states).
Let also $\mathrm{FA}_S = (Q, Q \smallsetminus \A, Q_f, q_0, \rules)$.  All
these automata only differ in their final state sets.

According to the automata theory, for each $\mathrm{FA}_i$ there exists an
equivalent regular expression $e_i$ over $Q\a$.  Since $Q \subset \A \cup N$,
each $e_i$ is a valid regular expression over $(\A \cup N)\a$ as well.	The
same applies to $e_S$, equivalent to $\mathrm{FA}_S$.

Now, let $R = \bigcup_i \{ n_i \rightarrow \langle e_i \rangle \} \:\cup\: \{
S \rightarrow \langle e_S \rangle \}$.  The resulting grammar $G = (\A, N, S,
R)$ is a regular \foo tree grammar, which recognises the language of automaton
$\auto$.
\end{proof}

\subsection{Vertical concatenation}
\label{sect:vert-conc}

Although the two tree operators, concatenation and encapsulation, are
sufficient to build all possible \foo trees out of basic elements, it may
occasionally be useful to link trees at places other than their roots.  For
example, a tree can be built from root downwards by repeatedly attaching other
trees at its leaves.  In tree automata theory this type of tree linking is
done by replacing a symbol in the first tree by the root of the second tree
(Figure~\ref{fig:conc-var}).  This is readily transferable to \foo trees.
\begin{figure}[htb]
\centering
\setlength{\unitlength}{0.00087489in}
\begingroup\makeatletter\ifx\SetFigFont\undefined%
\gdef\SetFigFont#1#2#3#4#5{%
  \reset@font\fontsize{#1}{#2pt}%
  \fontfamily{#3}\fontseries{#4}\fontshape{#5}%
  \selectfont}%
\fi\endgroup%
{\renewcommand{\dashlinestretch}{30}
\begin{picture}(5175,1530)(0,-10)
\put(2475,1035){\ellipse{254}{254}}
\put(2700,630){\ellipse{254}{254}}
\put(2205,630){\ellipse{254}{254}}
\path(2410,914)(2285,727)
\path(2520,900)(2610,720)
\put(2475,990){\makebox(0,0)[b]{\smash{{{\SetFigFont{12}{14.4}{\sfdefault}{\mddefault}{\updefault}f}}}}}
\put(2700,585){\makebox(0,0)[b]{\smash{{{\SetFigFont{12}{14.4}{\sfdefault}{\mddefault}{\updefault}h}}}}}
\put(2205,585){\makebox(0,0)[b]{\smash{{{\SetFigFont{12}{14.4}{\sfdefault}{\mddefault}{\updefault}g}}}}}
\put(4410,1350){\ellipse{254}{254}}
\put(4815,945){\ellipse{254}{254}}
\put(4410,945){\ellipse{254}{254}}
\put(4005,945){\ellipse{254}{254}}
\put(5040,540){\ellipse{254}{254}}
\put(4590,540){\ellipse{254}{254}}
\put(4815,135){\ellipse{254}{254}}
\put(4365,135){\ellipse{254}{254}}
\path(4320,1260)(4095,1035)
\path(4410,1215)(4410,1080)
\path(4500,1260)(4725,1035)
\path(4860,810)(4950,630)
\path(4770,810)(4670,637)
\path(4635,405)(4725,225)
\path(4545,405)(4445,232)
\put(4410,1305){\makebox(0,0)[b]{\smash{{{\SetFigFont{12}{14.4}{\sfdefault}{\mddefault}{\updefault}a}}}}}
\put(4005,900){\makebox(0,0)[b]{\smash{{{\SetFigFont{12}{14.4}{\sfdefault}{\mddefault}{\updefault}b}}}}}
\put(4410,900){\makebox(0,0)[b]{\smash{{{\SetFigFont{12}{14.4}{\sfdefault}{\mddefault}{\updefault}c}}}}}
\put(4815,900){\makebox(0,0)[b]{\smash{{{\SetFigFont{12}{14.4}{\sfdefault}{\mddefault}{\updefault}d}}}}}
\put(5040,495){\makebox(0,0)[b]{\smash{{{\SetFigFont{12}{14.4}{\sfdefault}{\mddefault}{\updefault}e}}}}}
\put(4590,495){\makebox(0,0)[b]{\smash{{{\SetFigFont{12}{14.4}{\sfdefault}{\mddefault}{\updefault}f}}}}}
\put(4815,90){\makebox(0,0)[b]{\smash{{{\SetFigFont{12}{14.4}{\sfdefault}{\mddefault}{\updefault}h}}}}}
\put(4365,90){\makebox(0,0)[b]{\smash{{{\SetFigFont{12}{14.4}{\sfdefault}{\mddefault}{\updefault}g}}}}}
\put(540,1350){\ellipse{254}{254}}
\put(540,945){\ellipse{254}{254}}
\put(135,945){\ellipse{254}{254}}
\put(945,945){\ellipse{254}{254}}
\put(1170,540){\ellipse{254}{254}}
\put(720,540){\ellipse{254}{254}}
\path(450,1260)(225,1035)
\path(630,1260)(855,1035)
\path(540,1215)(540,1080)
\path(990,810)(1080,630)
\path(900,810)(800,637)
\put(540,1305){\makebox(0,0)[b]{\smash{{{\SetFigFont{12}{14.4}{\sfdefault}{\mddefault}{\updefault}a}}}}}
\put(135,900){\makebox(0,0)[b]{\smash{{{\SetFigFont{12}{14.4}{\sfdefault}{\mddefault}{\updefault}b}}}}}
\put(540,900){\makebox(0,0)[b]{\smash{{{\SetFigFont{12}{14.4}{\sfdefault}{\mddefault}{\updefault}c}}}}}
\put(945,900){\makebox(0,0)[b]{\smash{{{\SetFigFont{12}{14.4}{\sfdefault}{\mddefault}{\updefault}d}}}}}
\put(1170,495){\makebox(0,0)[b]{\smash{{{\SetFigFont{12}{14.4}{\sfdefault}{\mddefault}{\updefault}e}}}}}
\put(720,495){\makebox(0,0)[b]{\smash{{{\SetFigFont{12}{14.4}{\sfdefault}{\mddefault}{\updefault}$x$}}}}}
\put(1665,810){\makebox(0,0)[b]{\smash{{{\SetFigFont{20}{24.0}{\familydefault}{\mddefault}{\updefault}$\cdot_x$}}}}}
\put(3285,810){\makebox(0,0)[b]{\smash{{{\SetFigFont{20}{24.0}{\familydefault}{\mddefault}{\updefault}=}}}}}
\end{picture}
}
\caption{Concatenation through variable $x$}
\label{fig:conc-var}
\end{figure}
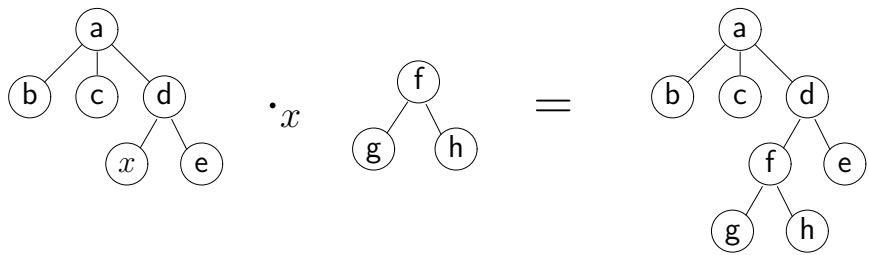
\begin{definition}
Vertical concatenation of two trees $u, v \in T(\A)$ through symbol $x \in
\A$, denoted $u \cdot_x v$, is the tree derived from $u$ by replacing all
occurrences of $x$ in it with $v$.
\end{definition}
Vertical concatenation can also be defined recursively as follows:
\begin{align*}
\langle\rangle \cdot_x t & = \langle\rangle
	& (u\cdot v)\cdot_x t & = (u\cdot_x t)\cdot(v\cdot_x t) \\
\langle a \rangle \cdot_x t & = \langle a \rangle,\, \text{ for } a\neq x
	& \langle u \rangle \cdot_x t & = \langle u\cdot_x t \rangle \\
\langle x \rangle \cdot_x t & = \langle t \rangle & &
	& \text{where } a, x \in \A;\; u, v, t \in T(\A).
\end{align*}
Note that vertical concatenations have no neutral element, because they always
insert at least a pair of angle brackets: $\langle axb \rangle \cdot_x \langle
x \rangle = \langle a\langle x\rangle b\rangle$.  By analogy with tree
automata theory, symbols which are used for vertical concatenation will be
called \emph{variables} to help us distinguish them from other symbols in the
alphabet.  Normally, symbols which can occur in the actual trees being
processed are not used as variables; the latter are chosen from some disjoint
set.

\subsection{Regular tree expressions}

Combination of string regular expressions and classical tree regular
expressions yields the following definition.
\begin{definition}
The set $\mathit{RSTE}(\A)$ of regular \foo tree expressions over $\A$ is
defined as follows:
\begin{itemize}
\item $\langle\rangle \in \mathit{RSTE}(\A)$;
\item $\A \subset \mathit{RSTE}(\A)$;
\item if $\:r_1, r_2 \in \mathit{RSTE}(\A),\:$ then $\:r_1|r_2 \in
	\mathit{RSTE}(\A)$\hfill (union);
\item if $\:r_1, r_2 \in \mathit{RSTE}(\A),\:$ then $\:r_1\cdot r_2 \in
	\mathit{RSTE}(\A)$\hfill (horizontal concatenation);
\item if $\:r_1, r_2 \in \mathit{RSTE}(\A)\:$ and $\:x \in \A,\:$ then
	$\:r_1\cdot_x r_2 \in \mathit{RSTE}(\A)$\hfill (vertical
	concatenation);
\item if $\:r \in \mathit{RSTE}(\A),\:$ then $\:r\a \in
	\mathit{RSTE}(\A)$\hfill (horizontal iteration);
\item if $\:r \in \mathit{RSTE}(\A)\:$ and $\:x \in \A,\:$ then
	$\:r\viter{x} \in \mathit{RSTE}(\A)$\hfill (vertical
	iteration).
\end{itemize}
Note that encapsulation, not mentioned above explicitly, is also a regular
operator: $\langle r \rangle = \langle x \rangle \cdot_x r$ for $x \in \A$.
\end{definition}

The language described by an RSTE $r$ is defined analogously to the classical
theories and is denoted as $\reglang{r}$.  Regular \foo tree expressions have
the same expressive power as regular \foo tree grammars and finite automata.

\subsubsection{Regular expression matching and substitution}

Regular expressions play a major role in many text processing tools, such as
`sed', `awk', `perl', typically found on Unix systems~\cite{sed-awk}.  They
allow matching and selecting pieces of text that can then be re-combined to
produce desired results.  First, the input text is matched against a regular
expression.  When a match is found, sub-expressions of that regular expression
are associated with the corresponding fragments of text that they match.
These fragments can then be extracted simply by referring to the desired
sub-expressions.

In the traditional tools, regular expressions are represented in a
parenthesised infix form.  Round brackets in expressions play a dual role:
they group regular operators and also identify sub-expressions that will be
used for text extraction.  These sub-expressions are then referred to by their
numbers (counted left-to-right according to their opening brackets).  For
example, expression `\symb{$($press$|$push$|$hit$|$strike$)$ space
$($key$|$bar$)$}', applied to the phrase `\symb{push space bar}', would result
in a successful match, selecting `\symb{push}' and `\symb{bar}' as fragments 1
and 2, respectively.

Below are a few examples of \foo tree regular expressions.  We assume that the
variables used in the expressions cannot occur in actual trees; in other
words, variables (denoted below as $X$, $Y$, $Z$) are unique symbols, added to
the input alphabet $\A$.  In our notation, operators have the following
priorities:
\begin{center}
\begin{tabular}{cc@{}l}
$\a$	& $\viter{x}$		& \quad\quad(highest priority)	\\
$\cdot$	& $\cdot_x$		&				\\
\multicolumn{2}{c}{\:$|$}	& \quad\quad(lowest priority)
\end{tabular}
\end{center}
We also use the symbol $\alpha$ as a convenient shorthand for the union of all
single symbol trees without variables.

The expression $(\alpha | \langle X\rangle)\a \,\viter{X}$ matches any tree.
This expression will be denoted as `$\anytree$', assuming that the variable
$X$ is not used anywhere in the expression that contains $\anytree$, as it may
cause unwanted interference.  In other words, the scope of $X$ in $\anytree$
is restricted to $\anytree$.  An expression that matches any tree with some
variable(s) will also be handy, e.g.: $\anytree_{XY} = (\alpha | \langle
X\rangle | \langle Y\rangle | \langle Z\rangle)\a \,\viter{Z}$, where $Z$ has
a restricted scope.

\begin{figure}[htb]
$$
\parbox{0.35\textwidth}{\setlength{\unitlength}{0.00087489in}
\begingroup\makeatletter\ifx\SetFigFont\undefined%
\gdef\SetFigFont#1#2#3#4#5{%
  \reset@font\fontsize{#1}{#2pt}%
  \fontfamily{#3}\fontseries{#4}\fontshape{#5}%
  \selectfont}%
\fi\endgroup%
{\renewcommand{\dashlinestretch}{30}
\begin{picture}(2805,2543)(0,-10)
\put(372,368){\makebox(0,0)[b]{\smash{{{\SetFigFont{12}{14.4}{\sfdefault}{\mddefault}{\updefault}...}}}}}
\put(1452,368){\makebox(0,0)[b]{\smash{{{\SetFigFont{12}{14.4}{\sfdefault}{\mddefault}{\updefault}...}}}}}
\put(912,368){\makebox(0,0)[b]{\smash{{{\SetFigFont{12}{14.4}{\sfdefault}{\mddefault}{\updefault}...}}}}}
\put(628.810,2198.259){\arc{514.465}{0.9569}{3.1989}}
\put(1195.190,2198.259){\arc{514.467}{6.2258}{8.4679}}
\put(912.000,2190.500){\arc{675.000}{0.6435}{2.4981}}
\put(912,503){\ellipse{540}{990}}
\path(912,1943)(912,1538)
\path(822,2483)(462,2123)
\path(1002,2483)(1362,2123)
\path(732,1358)(12,953)
\path(1092,1358)(1812,953)
\path(912,1313)(912,953)
\path(822,1313)(462,953)
\path(1002,1313)(1362,953)
\path(957,728)(1137,323)
\path(867,728)(687,323)
\path(507,728)(552,323)
\path(1407,728)(1632,323)
\path(1317,728)(1272,323)
\path(1407,143)(1047,233)
\blacken\path(1170.693,233.000)(1047.000,233.000)(1156.141,174.791)(1128.492,212.627)(1170.693,233.000)
\path(417,728)(192,323)
\put(912,2483){\makebox(0,0)[b]{\smash{{{\SetFigFont{12}{14.4}{\sfdefault}{\mddefault}{\updefault}*}}}}}
\put(912,2168){\makebox(0,0)[b]{\smash{{{\SetFigFont{17}{20.4}{\sfdefault}{\mddefault}{\updefault}.  .  .}}}}}
\put(912,1313){\makebox(0,0)[b]{\smash{{{\SetFigFont{12}{14.4}{\sfdefault}{\mddefault}{\updefault}*}}}}}
\put(327,998){\makebox(0,0)[b]{\smash{{{\SetFigFont{12}{14.4}{\sfdefault}{\mddefault}{\updefault}...}}}}}
\put(1497,998){\makebox(0,0)[b]{\smash{{{\SetFigFont{12}{14.4}{\sfdefault}{\mddefault}{\updefault}...}}}}}
\put(462,773){\makebox(0,0)[b]{\smash{{{\SetFigFont{12}{14.4}{\sfdefault}{\mddefault}{\updefault}left}}}}}
\put(912,728){\makebox(0,0)[b]{\smash{{{\SetFigFont{12}{14.4}{\sfdefault}{\mddefault}{\updefault}*}}}}}
\put(1452,53){\makebox(0,0)[lb]{\smash{{{\SetFigFont{12}{14.4}{\familydefault}{\mddefault}{\updefault}selected sub-tree}}}}}
\put(1407,773){\makebox(0,0)[b]{\smash{{{\SetFigFont{12}{14.4}{\sfdefault}{\mddefault}{\updefault}right}}}}}
\end{picture}
}}
\anytree_X \cdot_X (\anytree \cdot \langle\langle \symb{left} \rangle \cdot
\langle \anytree \rangle\a \rangle \cdot \langle
\,\underbrace{(\anytree)}_{\makebox[0pt]{selection}}\, \rangle \cdot \langle
\langle \symb{right} \rangle \cdot \langle \anytree \rangle\a \rangle \cdot
\anytree)
$$
\caption{Selecting a subtree with a regular expression}
\label{fig:regexp_example}
\end{figure}
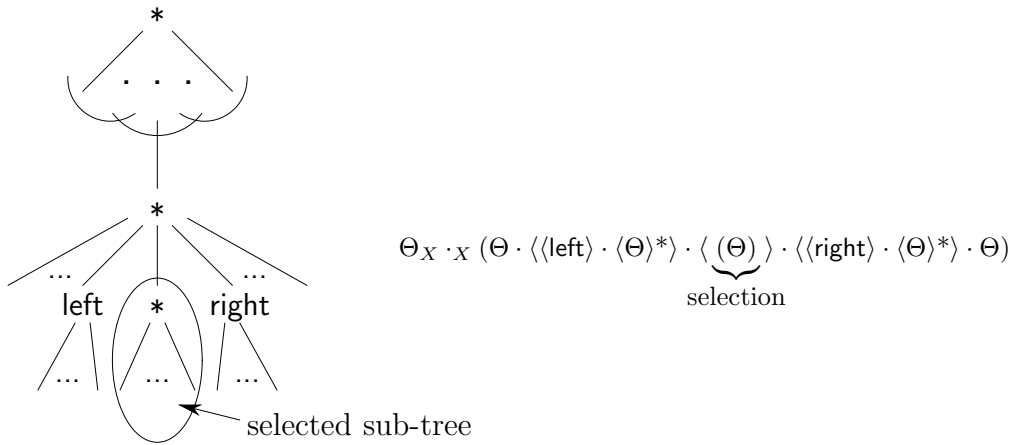

Figure~\ref{fig:regexp_example} shows a regular expression that selects a
subtree located tightly between two subtrees labelled \symb{left} and
\symb{right} (in this order) somewhere below the root of the input tree.
The requested subtree is selected by the second pair of round brackets.  The
first pair is needed for grouping operators and (as a side effect) selects the
parent tree of the one we are looking for.

What if the same regular sub-expression matches more than one fragment?  This
is largely an implementation issue and is up to a particular application.  Let
us consider potential solutions.

To begin with, we need to separate two cases when an expression can match more
than one fragment: ambiguity and multiplicity.  \emph{Ambiguity} happens when
a sub-expression can match one out of several alternatives, for example, in an
expression like $r\a \cdot (r) \cdot r\a$.  This is the same as $r\a$---a
concatenation of zero or more trees described by $r$.  The $(r)$ in the middle
can match any tree in the sequence---the first one, the last one, or a tree
somewhere in between.  \emph{Multiplicity} happens when the same
sub-expression matches several fragments simultaneously, as in $(r)\a$.  This
describes the same sequence as the previous example, but this time $(r)$
matches \emph{all} the trees that compose that sequence.  By definition of
iteration, there are in fact infinitely many copies of $r$ in that expression,
each copy matching at most one tree; when these copies are represented in the
formula by a single sub-expression, it happens to match all those trees
simultaneously.

Ambiguity is traditionally solved by imposing longest-match or shortest-match
rules, or by forbidding ambiguous regular expressions.  In case of
multiplicity, usually only the first or the last matching fragment is
selected.  However, as opposed to the text case, a tree processing application
has the benefit of hierarchical structure: several fragments can be combined
into one object simply by adding another level of hierarchy.  Thus, a regular
sub-expression that can potentially exhibit multiplicity may be associated
with a tree whose subtrees are the fragments matched.

\section{Potential applications of the \foo tree automata theory}
\label{sect:app_scenarios}

\begin{figure}[htb]
\centering
\setlength{\unitlength}{0.00087489in}
\begingroup\makeatletter\ifx\SetFigFont\undefined%
\gdef\SetFigFont#1#2#3#4#5{%
  \reset@font\fontsize{#1}{#2pt}%
  \fontfamily{#3}\fontseries{#4}\fontshape{#5}%
  \selectfont}%
\fi\endgroup%
{\renewcommand{\dashlinestretch}{30}
\begin{picture}(6510,2010)(0,-10)
\put(126,1078){\blacken\ellipse{80}{80}}
\put(126,1078){\ellipse{80}{80}}
\put(285,1078){\blacken\ellipse{80}{80}}
\put(285,1078){\ellipse{80}{80}}
\put(126,918){\blacken\ellipse{80}{80}}
\put(126,918){\ellipse{80}{80}}
\put(285,918){\blacken\ellipse{80}{80}}
\put(285,918){\ellipse{80}{80}}
\put(444,918){\blacken\ellipse{80}{80}}
\put(444,918){\ellipse{80}{80}}
\put(206,1236){\blacken\ellipse{80}{80}}
\put(206,1236){\ellipse{80}{80}}
\path(126,1078)(206,1236)(285,1078)(444,918)
\path(126,918)(285,1078)(285,918)
\put(285,1018){\ellipse{554}{908}}
\put(285,404){\makebox(0,0)[b]{\smash{{{\SetFigFont{12}{14.4}{\familydefault}{\mddefault}{\updefault}input}}}}}
\put(6264,1236){\blacken\ellipse{80}{80}}
\put(6264,1236){\ellipse{80}{80}}
\put(6106,1078){\blacken\ellipse{80}{80}}
\put(6106,1078){\ellipse{80}{80}}
\put(6264,1078){\blacken\ellipse{80}{80}}
\put(6264,1078){\ellipse{80}{80}}
\put(6422,1078){\blacken\ellipse{80}{80}}
\put(6422,1078){\ellipse{80}{80}}
\path(6106,1078)(6264,1236)(6264,1078)
\path(6264,1236)(6422,1078)
\put(6027,918){\blacken\ellipse{80}{80}}
\put(6027,918){\ellipse{80}{80}}
\put(6185,918){\blacken\ellipse{80}{80}}
\put(6185,918){\ellipse{80}{80}}
\put(6344,918){\blacken\ellipse{80}{80}}
\put(6344,918){\ellipse{80}{80}}
\put(6225,1018){\ellipse{554}{908}}
\path(6027,918)(6106,1078)(6185,918)
\path(6264,1078)(6344,918)
\put(6225,404){\makebox(0,0)[b]{\smash{{{\SetFigFont{12}{14.4}{\familydefault}{\mddefault}{\updefault}output}}}}}
\put(3255,1156){\blacken\ellipse{80}{80}}
\put(3255,1156){\ellipse{80}{80}}
\put(3097,998){\blacken\ellipse{80}{80}}
\put(3097,998){\ellipse{80}{80}}
\put(3255,998){\blacken\ellipse{80}{80}}
\put(3255,998){\ellipse{80}{80}}
\put(3413,998){\blacken\ellipse{80}{80}}
\put(3413,998){\ellipse{80}{80}}
\path(3097,998)(3255,1156)(3255,998)
\path(3255,1156)(3413,998)
\put(3255,1040){\ellipse{476}{476}}
\put(3175,1711){\blacken\ellipse{80}{80}}
\put(3175,1711){\ellipse{80}{80}}
\put(3335,1711){\blacken\ellipse{80}{80}}
\put(3335,1711){\ellipse{80}{80}}
\put(3255,1869){\blacken\ellipse{80}{80}}
\put(3255,1869){\ellipse{80}{80}}
\put(3255,1790){\ellipse{396}{396}}
\path(3175,1711)(3255,1869)(3335,1711)
\put(3255,245){\blacken\ellipse{80}{80}}
\put(3255,245){\ellipse{80}{80}}
\put(3255,87){\blacken\ellipse{80}{80}}
\put(3255,87){\ellipse{80}{80}}
\put(3255,404){\blacken\ellipse{80}{80}}
\put(3255,404){\ellipse{80}{80}}
\put(3255,245){\ellipse{476}{476}}
\path(3255,404)(3255,245)(3255,87)
\drawline(3255,245)(3255,245)
\path(641,1038)(879,1038)
\blacken\path(773.400,1011.605)(879.000,1038.000)(773.400,1064.395)(805.080,1038.000)(773.400,1011.605)
\path(5631,1038)(5869,1038)
\blacken\path(5763.400,1011.605)(5869.000,1038.000)(5763.400,1064.395)(5795.080,1038.000)(5763.400,1011.605)
\path(2622,1038)(2938,1038)
\blacken\path(2832.400,1011.605)(2938.000,1038.000)(2832.400,1064.395)(2864.080,1038.000)(2832.400,1011.605)
\path(3572,1038)(3888,1038)
\blacken\path(3782.400,1011.605)(3888.000,1038.000)(3782.400,1064.395)(3814.080,1038.000)(3782.400,1011.605)
\path(2622,1236)(3017,1631)
\blacken\path(2960.994,1537.665)(3017.000,1631.000)(2923.665,1574.994)(2964.731,1578.731)(2960.994,1537.665)
\path(2622,840)(3017,443)
\blacken\path(2923.807,499.242)(3017.000,443.000)(2961.229,536.476)(2964.863,495.401)(2923.807,499.242)
\blacken\path(3794.665,1292.006)(3888.000,1236.000)(3831.994,1329.335)(3835.731,1288.269)(3794.665,1292.006)
\path(3888,1236)(3493,1631)
\blacken\path(3832.229,746.524)(3888.000,840.000)(3794.807,783.758)(3835.863,787.599)(3832.229,746.524)
\path(3888,840)(3493,443)
\path(958,1394)(2542,1394)(2542,681)
	(958,681)(958,1394)
\path(3968,1394)(5552,1394)(5552,681)
	(3968,681)(3968,1394)
\put(1750,998){\makebox(0,0)[b]{\smash{{{\SetFigFont{10}{12.0}{\familydefault}{\mddefault}{\updefault}expression matcher}}}}}
\put(1750,800){\makebox(0,0)[b]{\smash{{{\SetFigFont{10}{12.0}{\familydefault}{\mddefault}{\updefault}(automaton)}}}}}
\put(4760,998){\makebox(0,0)[b]{\smash{{{\SetFigFont{10}{12.0}{\familydefault}{\mddefault}{\updefault}(using concatenation}}}}}
\put(4760,800){\makebox(0,0)[b]{\smash{{{\SetFigFont{10}{12.0}{\familydefault}{\mddefault}{\updefault}and encapsulation)}}}}}
\put(4760,1196){\makebox(0,0)[b]{\smash{{{\SetFigFont{10}{12.0}{\familydefault}{\mddefault}{\updefault}Tree formula}}}}}
\put(1750,1196){\makebox(0,0)[b]{\smash{{{\SetFigFont{10}{12.0}{\familydefault}{\mddefault}{\updefault}Tree regular}}}}}
\end{picture}
}
\caption{Tree manipulation using regular \foo tree expressions}
\label{fig:app1}
\end{figure}
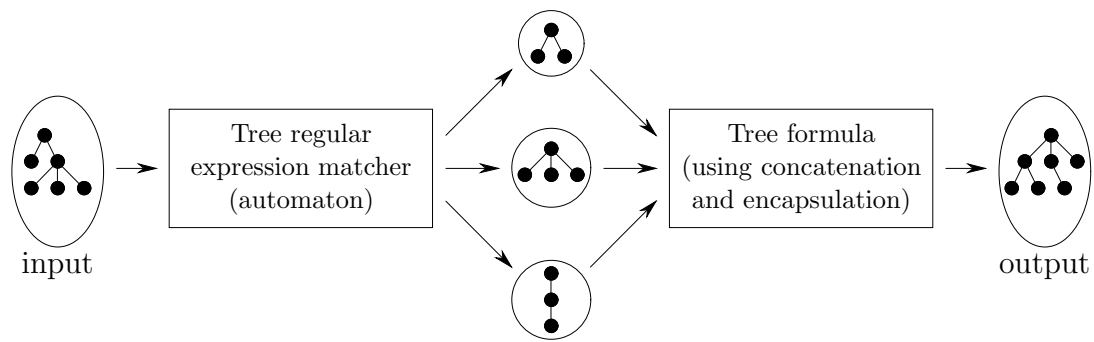

The data model described in this paper enables processing of structured
information in the same way as it is done today with non-structured textual
data.  In many text processing utilities and applications, data records are
matched against one or more string regular expressions.  Some parts of these
expressions are marked.  When a match occurs, each marked part is associated
with the piece of text that matched that part within the regular expression.
These pieces of text, extracted from the data record, are then used according
to the application's needs.  They can be combined together and with other
pieces (e.g., string literals) using concatenation, or they can be processed
further as simple strings.  The same basic processing scenario applies to our
hierarchical data model (as illustrated in Figure~\ref{fig:app1}).

An information processing application which uses the proposed data model needs
to implement a set of operations on trees that can be used in formulae (the
right box in Figure~\ref{fig:app1}).  These operations will likely include the
basic tree operators (encapsulation, concatenation) and some traditional
string functions (character translation, table lookup, etc.)  Note that many
of these functions can be expressed using regular expression matching,
concatenation, and string literals---thus, they do not extend the model, but
merely act as convenient shortcuts.  Tree processing primitives (such as
sorting of children) can also be included in the application's operational
model.

The application may also employ some execution model which specifies how (in
what order, on what conditions) tree operators are invoked.  For example, the
execution model can provide variables for storing trees.  In one of envisaged
scenarios, a tree transformation utility may treat the transformation it is
performing as a sequence of rules, each consisting of a condition and an
action.  The utility would execute rules sequentially by checking the
condition and, if necessary, doing the corresponding action.  Actions would
normally consist of concatenating trees taken from variables and constant
expressions, and storing them back into variables.

As suggested by the practices of using string regular expressions in text
manipulation tools, in such scenarios regular expressions on trees can play a
dual role:
\begin{itemize}
\item to serve as a rule condition by telling whether a particular tree
matches a pattern or not;
\item to ``extract'' parts of a tree.
\end{itemize}

Types of applications that can benefit from this data model include:
\begin{itemize}
\item stand-alone processing tools (generic or specialised), such as HTML or
XML processors or MARC (Machine Readable Catalogue~\cite{iso2709}) converters;
\item programming languages that include hierarchical data types;
\item information retrieval;
\item query languages.
\end{itemize}

\section{Conclusions}
\label{sect:conclusions}

The ``\foo tree'' data model introduced in this paper provides a simple
algebraic notion for the hierarchical information structures.  The model is
based on two classical works: theory of automata and formal languages; and
tree automata theory.  The classical notions are extended and combined
together to provide a powerful solution for today's information processing
needs.  The resulting model combines the expressive power of its parents:
strings, trees, finite automata, and regular languages defined by the
classical theories can be expressed in the proposed model.  It is shown that
many of the properties of classical models apply here as well.  Therefore, it
is reasonable to expect that even those techniques that are used in classical
theories but not considered explicitly in this paper, can be formulated and
re-used in terms of the proposed model.

The most important conclusion is that processing of tree-like hierarchical
data can benefit from the power of regular expressions in the same way that
simple text processing has benefited from them to date.  It is shown that
regular expression matching on trees can be done by finite tree automata,
which fit into linear memory and space constraints.

The formal description of the proposed data model supplies a basis for
building custom, problem-oriented, as well as generic solutions for data
retrieval and processing.  These solutions can build upon additional operators
introduced on trees.

The model offered is simple, consisting of only three operators:
concatenation, encapsulation, and regular expression matching.  The regular
algebra contains five operators.  Many processing functions, including
frequently used convenience operators (such as, for example, extraction of
$n$-th subtree) can be expressed using the basic set of operators without the
need to extend the formal model.  This puts the proposed model in a favourable
position with respect to existing solutions, which are often over-complicated.
Currently, different bits of tree manipulation functionality (such as
execution model, string functions, tree operators, extraction of sub-records)
are often bundled together and depend on each other.  This makes existing
models cumbersome and inflexible.  Also, if, for instance, such model is
incorporated into a programming language to provide it with a ``tree'' data
type, that language is likely to already offer string and numeric processing.
This results in duplicate functionality.

Finally, the model described in this paper provides a unified representation
of both information trees and character strings, which makes it suitable as a
single data type for processing of both structure and content of hierarchical
documents.  A simple scenario is presented showing how this model can be used
in information processing applications.  This scenario is patterned
after the current usage of traditional regular expressions in unstructured
text processing.

Further research can be centred along the following lines:
\begin{itemize}

\item \emph{Use of non-string data types} (e.g., numeric and Boolean), both in
the data being processed and in the operational model.  The present data model
is capable of processing, say, numeric data if the application's operational
model supports it (i.e., it contains arithmetic and conversion operators).
However, feeding numeric data back to the regular expression matching engine
is not obvious.  This may be needed, for example, to extract a subtree by its
number, where this number is determined dynamically---much like indexing an
array by a variable.  Of some relevance here is research on XML that has been
investigating how data typing can be incorporated into XML
schemata~\cite{xml-schema}.

\item \emph{Presentation of regular expressions}.  Tree regular expressions
written in infix form (as in Figure~\vref{fig:regexp_example}) look more
complex and may be more difficult to compose than string regular expressions.
On the other hand, this notation is also compact and close to the conventional
syntax.  Another possible representation may be that of the regular tree
grammar, which has the advantage of being similar to SGML/XML DTD format.

\end{itemize}

\appendix
\section{Proof of equivalence of deterministic and generalised
tree automata}
\label{app:equiv}

This Appendix contains the proof of Theorem~\ref{stmt:dg-equiv} on
page~\pageref{stmt:dg-equiv} that states that any generalised
non-deterministic finite \foo tree automaton (GNFSTA) with ``pure'' states has
an equivalent deterministic automaton (DFSTA).  The basic principle behind
this proof is the same as the one used in the classical automata theory to
prove the equivalence of deterministic and non-deterministic automata, namely
subset construction.  For an arbitrary GNFSTA, we build an equivalent DFSTA
whose state set is the set of all subsets of the original GNFSTA's state set.
A brief sketch of this proof is presented in Section~\ref{sect:equiv}, where
the theorem is introduced.

The requirement that the original GNFSTA be an automaton with ``pure''
states allows us to distinguish original tree's input symbols from
states generated during the automaton's run.  As shown in
Section~\ref{sect:pure}, any GNFSTA can be converted to an automaton with pure
states.

In the next section, we introduce two auxiliary items: a map $\Gamma$ and a
relation $\inst$, which will be used in the proof.  Then, the following
section proves a lemma which incorporates most of the work.  After that, the
validity of the main result of Theorem~\ref{stmt:dg-equiv} follows almost
immediately from the lemma.

\subsection{Preliminaries}

Let $\auto$ be a GNFSTA with pure states over $\A$: $\auto = (\A, Q, Q_f, Q_0,
\rules, \gamma)$.  As suggested above (and in the sketch of the proof in
Section~\ref{sect:equiv}), we want to construct a suitable DFSTA $\auto' =
(\A', Q', Q'_f, q'_0, \rules')$, where $Q' = 2^Q$, and prove their
equivalence.  First, however, a few auxiliary concepts need to be introduced.

Let $T = T(Q \cup \{/\}), \quad T' = T(2^Q \cup \{/\})$.  These sets contain
trees on which the move relations $\moverel$ and $\Moverel[\auto']{}$ are
defined.  That is, all intermediate trees which compose runs of $\auto$ and
$\auto'$ belong to $T$ and $T'$, respectively.

Because $\auto$ has pure states, its ``vertical move'' function $\gamma$
produces empty set on all symbols from $\A$.  Let us redefine $\gamma$ on
$\A$ and then extend it to $2^Q$ as follows:
\begin{align*}
\gamma(a) &= \{a\} \quad \text{for all}\; a \in \A; \\
\gamma(\{q_1,\dots, q_n\}) &= \gamma(q_1) \cup \dots \cup \gamma(q_n).
\end{align*}
We now define function $\Gamma: T' \rightarrow T'$, which applies $\gamma$ to
all the input symbols in intermediate trees for $\auto'$.  That is, if a label
in the tree is yet ``untouched'' by the automaton (does not contain /),
$\gamma$ is applied to all symbols of the label.  If a label is partly
processed, $\gamma$ is only applied to the part on the right hand side of the
slash (/).  Remembering that symbols in $T'$ are \emph{sets} of symbols of
$T$, we define $\Gamma$ using string representations of trees from $T'$ as
follows.  Let $a$ denote a single state of $T'$: $a \in 2^Q$.  Let $s$ denote
an arbitrary sub-string of a tree from $T'$, which does not start with /: $s
\in (2^Q \cup \{\langle, \rangle, /\})\a,\, s \neq /x$.  Then, let
\begin{align*}
\Gamma(\varepsilon) &= \varepsilon
	& \Gamma(\langle s) &= \langle\Gamma(s)
	& \Gamma(as) &= \gamma(a)\Gamma(s) \\
&
	& \Gamma(\rangle s) &= {}{\rangle}\Gamma(s)
	& \Gamma(a/s) &= a/\Gamma(s)
\end{align*}
Note that by this definition, $\Gamma(xy) = \Gamma(x)\Gamma(y)$, if $y$ does
not start with a slash.  Also, if a tree $t'$ is not just an arbitrary tree from
$T'$, but an intermediate stage of the automaton $\auto'$, each / in $t'$ must
contain a state from $Q'$ immediately on its left.  This implies, in
particular, that for any representation $t' = l'\langle s'\rangle r'$, it is
true that $\Gamma(t') = \Gamma(l')\langle\Gamma(s')\rangle\Gamma(r')$.

Another thing that will be needed for the proof is a relation between $T$ and
$T'$, which is induced by the simple ``belongs to'' relation between a set and
its element.  Let $t \in T$ and $t' \in T'$.  We say that $t$ is an
\emph{instance} of $t'$ (denoted $t \inst t'$), if:
\begin{enumerate}
\renewcommand{\labelenumi}{(\alph{enumi})}
\item $t$ and $t'$ have the same structure, i.e.\ symbols $\langle$,
$\rangle$, and / occupy the same positions in both trees; and
\item each $Q$-symbol in $t$ belongs to the set of symbols in the same
position in $t'$.
\end{enumerate}
Or, more formally,
%\begin{align*}
%& \varepsilon\inst\varepsilon \\
%as \inst a's'	& \:\Longleftrightarrow\: a \in a' \:\text{and}\: s \inst s'
%	& \text{for any}\; & a \in Q,\, a' \in 2^Q, \\
%{\langle} s \inst {\langle} s'	& \:\Longleftrightarrow\: s \inst s'
%	& & s \in (Q \cup \{\langle, \rangle, /\})\a, \\
%{\rangle} s \inst {\rangle} s'	& \:\Longleftrightarrow\: s \inst s'
%	& & s' \in (2^Q \cup \{\langle, \rangle, /\})\a \\
%/s \inst /s'			& \:\Longleftrightarrow\: s \inst s'
%\end{align*}
\begin{align*}
& \varepsilon\inst\varepsilon \\
as \inst a's'	& \:\Longleftrightarrow\: a \in a' \:\text{and}\: s \inst s'
	& & \text{for any} \\
{\langle} s \inst {\langle} s'	& \:\Longleftrightarrow\: s \inst s'
	& & a \in Q,\, a' \in 2^Q, \\
{\rangle} s \inst {\rangle} s'	& \:\Longleftrightarrow\: s \inst s'
	& & s \in (Q \cup \{\langle, \rangle, /\})\a, \\
/s \inst /s'			& \:\Longleftrightarrow\: s \inst s'
	& & s' \in (2^Q \cup \{\langle, \rangle, /\})\a
\end{align*}
(where $\Longleftrightarrow$ denotes logical equivalence).
An immediate corollary is that $xy \inst x'y'$ is equivalent to $x \inst x',\,
y \inst y'$ for any strings $x,\,y,\,x',\,y'$ taken from the appropriate
string sets.

\subsection{Equivalence of DFSTA and GNFSTA}

Given a generalised non-deterministic finite \foo tree automaton $\auto = (\A,
Q, Q_f, Q_0, \rules, \gamma)$ with pure states, let us define DFSTA $\auto' =
(\A', Q', Q'_f, q'_0, \rules')$ as follows:
%\begin{eqnarray*}
%& &	\A' = \{ \{a\} \:|\: a \in \A \}; \quad\quad\quad\quad Q' = 2^Q;
%		\quad\quad\quad\quad q'_0 = Q_0; \\
%& &	Q'_f = \{ q' \in Q' \:|\: q' \cap Q_f \neq \emptyset \}; \\
%& &	\rules'(a', b') = c',\quad c' = \{ c \in Q \:|\: (a, b)\rightarrow c
%		\in \rules,\, a \in a',\, b \in \gamma(b') \}\quad \text{for
%		all}\; a', b' \in Q',
%\end{eqnarray*}
\begin{eqnarray*}
& &	\A' = \{ \{a\} \:|\: a \in \A \}; \quad\quad
		Q' = 2^Q; \quad\quad q'_0 = Q_0; \quad\quad
		Q'_f = \{ q' \in Q' \:|\: q' \cap Q_f \neq \emptyset \}; \\
& &	\rules'(a', b') = \{ c \in Q \:|\: (a, b)\rightarrow c
		\,\in\, \rules,\, a \in a',\, b \in \gamma(b') \}\quad
		\text{for all}\; a', b' \in Q',
\end{eqnarray*}
where $\gamma$ is understood in the extended sense.  As follows from the
definition of $\A'$, the tree monoids $T(\A)$ and $T(\A')$ are identical up to
renaming of symbols: symbols from $\A$ map to the corresponding single
element sets in $\A'$.	To prove the equivalence of $\auto$ and $\auto'$, all
we need to do is to prove that for any $t \in T(\A)$ and its counterpart $t'
\in T(\A')$, $t$ belongs to $L(\auto)$ if and only if $t'$ belongs to
$L(\auto')$.

To do this, we firstly consider the set of all intermediate trees, reachable
from $t$ in $n$ steps under $\moverel$, and, analogously, the set of all trees
reachable from $t'$ via $\Moverel[\auto']{n}$.	We shall prove that the first
set contains exactly all instances of $\Gamma$-images of trees from the second
set.

\begin{lemma}
\label{stmt:inst-of-steps}
Let $\steps{n}{\auto}(t)$ denote such sets: $\steps{n}{\auto}(t) = \{ v \in
T(Q_{\auto} \cup \{/\}) \:|\: t \Moverel{n} v \}$.  Then for any $v \in
\steps{n}{\auto}(t)$, $v$ is an instance of $\Gamma$-image of some $v' \in
\steps{n}{\auto'}(t')$ $(v \inst \Gamma(v'))$, and for any $v' \in
\steps{n}{\auto'}(t')$ all of the instances of $\Gamma(v')$ belong to
$\steps{n}{\auto}(t)$.
\end{lemma}

\begin{proof}
Induction on $n$.

\basis $n = 0$.  Then $\steps{0}{\auto}(t) = \{ t \}$,
$\steps{0}{\auto'}(t') = \{ t' \}$, $\Gamma(t') = t'$ (because $t'$ only
contains symbols from $\A'$), and $t \inst t'$ by the definition of $\inst$.

\induction  By the inductive hypothesis, the statement is assumed to be true
for $\steps{n}{}$.  We need to prove that:
\begin{enumerate}
\renewcommand{\theenumi}{(\roman{enumi})}
\renewcommand{\labelenumi}{\theenumi}
\item\label{item:vtovprime}
	for any $v \in \steps{n+1}{\auto}(t)$ there exists $v' \in
	\steps{n+1}{\auto'}(t')$ such that $v \inst \Gamma(v')$;
\item\label{item:vprimetov}
	for any $v' \in \steps{n+1}{\auto'}(t')$ and $v \in T(Q \cup \{/\})$,
	if $v \inst \Gamma(v')$, then $v \in \steps{n+1}{\auto}(t)$.
\end{enumerate}
Let us start by proving \ref{item:vtovprime}.  Since $v \in
\steps{n+1}{\auto}(t)$, there exists $u \in \steps{n}{\auto}(t)$ such that $u
\moverel v$.  Therefore, by the inductive hypothesis there exists $u' \in
\steps{n}{\auto'}(t')$ such that $u \inst \Gamma(u')$.	Consider all possible
moves from $u$ to $v$.
\begin{enumerate}
\renewcommand{\labelenumi}{(\alph{enumi})}
\item % initial state assignment
$u = l \langle s \rangle r,\: v = l \langle q_0/s \rangle r$, where $q_0 \in
Q_0$.

Since $u \inst \Gamma(u')$, we can write $u' = l' \langle s' \rangle
r'$, where $l \inst \Gamma(l')$, $s \inst \Gamma(s')$, $r \inst \Gamma(r')$,
and $s'$ does not contain /.  Consider now $v' = l' \langle Q_0/s' \rangle
r'$.  By the initial state assignment move of $\auto'$, $u'
\Moverel[\auto']{} v'$ (remembering that $q'_0 = Q_0$).  At the same time,
$\Gamma(v') = \Gamma(l') \langle Q_0/\Gamma(s') \rangle \Gamma(r')$; since
$q_0 \in Q_0$, we get that $v \inst \Gamma(v')$.

\item % horizontal move
$u = l \langle a/bs \rangle r,\: v = l \langle c/s \rangle r$, where $(a, b)
\rightarrow c \in \rules$.

Then, $u'$ can be represented as $u' = l' \langle
a'/b's' \rangle r'$, where $a \in a',\: b \in \gamma(b')$, and $l,s,r$ are
instances of $\Gamma(l'),\,\Gamma(s'),\,\Gamma(r')$ respectively.  From
definition of $\auto'$ it immediately follows that $\rules'(a',b') = c' \ni
c$.  Let $v' = l' \langle c'/s' \rangle r'$, then $u' \Moverel[\auto']{} v'$
and $v \inst \Gamma(v')$.

\item % vertical move
$u = l \langle q_1/ \rangle r,\: v = lq_2r$, where $q_1 \rightarrow q_2 \in
\gamma$ or, in other words, $q_2 \in \gamma(q_1)$.

Then, $u' = l' \langle
q'/ \rangle r'$, where $q_1 \in q',\: l \inst \Gamma(l'),\: r \inst
\Gamma(r')$.  Consider $v' = l'q'r'$; by the vertical move of $\auto'$, $u'
\Moverel[\auto']{} v'$.  Now, $q_2 \in \gamma(q_1) \subseteq \gamma(q')$.
Because $r$ cannot start with a /, neither can $r'$, therefore $\Gamma(v') =
\Gamma(l')\gamma(q')\Gamma(r')$, and consequently, $v \inst \Gamma(v')$.
\end{enumerate}

Let us now prove \ref{item:vprimetov}.	Since $v' \in
\steps{n+1}{\auto'}(t')$, there exists $u' \in \steps{n}{\auto'}(t')$ such
that $u' \Moverel[\auto']{} v'$.  Therefore, by the inductive hypothesis for
any $u \in T(Q \cup \{/\})$, $u \inst \Gamma(u')$ implies that $u \in
\steps{n}{\auto}(t)$.  We need to show that $v \inst \Gamma(v')$ implies $v
\in \steps{n+1}{\auto}(t)$.  Let us take an arbitrary $v$ such that $v \inst
\Gamma(v')$, and consider all possible moves from $u'$ to $v'$.

\begin{enumerate}
\renewcommand{\labelenumi}{(\alph{enumi})}
\item % initial state assignment
$u' = l' \langle s' \rangle r',\: v' = l' \langle q'_0/s' \rangle r'$.

Remembering that $q'_0 = Q_0$, and because $v \inst \Gamma(v') = \Gamma(l')
\langle Q_0/\Gamma(s') \rangle \Gamma(r')$, we can write $v$ as $v = l \langle
q_0/s \rangle r$, where $l,s,r$ are instances of $\Gamma(l'),\,
\Gamma(s'),\, \Gamma(r')$, and $q_0 \in Q_0$.  Let $u = l \langle s \rangle
r$.  By the properties of $\Gamma$ and the $\inst$-relation, $u \inst
\Gamma(u')$, therefore by the inductive hypothesis $u \in
\steps{n}{\auto}(t)$.  On the other hand, $u \moverel v$ by the initial state
assignment move of $\auto$.  Consequently, $v \in \steps{n+1}{\auto}(t)$.

\item % horizontal move
$u' = l' \langle a'/b's' \rangle r',\: v = l' \langle c'/s' \rangle r'$, and
$\rules'(a', b') = c'$.

We know that $v \inst \Gamma(v') = \Gamma(l') \langle
c'/\Gamma(s') \rangle \Gamma(r')$, which implies that $v = l \langle c/s
\rangle r$ for some $l, c, s, r$ such that $l,s,r$ are instances of
$\Gamma(l'),\, \Gamma(s'),\, \Gamma(r')$, and $c \in c'$.  Since $c \in
\rules'(a', b')$ and by definition of $\rules'$ (which we used to build our
$\auto'$), there exist $a \in a'$, $b \in \gamma(b')$, and rule $(a, b)
\rightarrow c \in \rules$.  Let $u = l \langle a/bs \rangle r$.  Using
definition and properties of $\Gamma$, we get $\Gamma(u') = \Gamma(l') \langle
a'/\gamma(b')\Gamma(s') \rangle \Gamma(r')$, so that $u \inst \Gamma(u')$.
Using the inductive hypothesis and the fact that $u \moverel v$ by the
horizontal move of $\auto$, we can observe that $v \in \steps{n+1}{\auto}(t)$.

\item % vertical move
$u' = l' \langle q'/ \rangle r',\: v' = l'q'r'$, and $q' \in Q'$.

Since $r'$
cannot start with a /, $v \inst \Gamma(v') = \Gamma(l')\gamma(q')\Gamma(r')$.
Then, $v = lq_2r$, where $l \inst \Gamma(l'),\: r \inst \Gamma(r')$, and $q_2
\in \gamma(q')$.  Because $\gamma$ of a set is the union of $\gamma$-images of
all elements of that set, there must be $q_1 \in q'$ such that $q_2 \in
\gamma(q_1)$.  Now let $u = l \langle q_1/ \rangle r$.	Because $\Gamma(u') =
\Gamma(l') \langle q'/ \rangle \Gamma(r')$ and $q_1 \in q'$, we get $u \inst
\Gamma(u')$ and use the inductive hypothesis.  Also, $q_2 \in \gamma(q_1)$
implies that $u \moverel v$ by the vertical move of $\auto$.  Therefore, $v
\in \steps{n+1}{\auto}(t)$.
\end{enumerate}
\vspace{-2ex}
\end{proof}

Now, it only remains to show that a final state is either reachable or
non-reachable simultaneously from both $t$ and $t'$.  In other words, we want
to prove that $\langle q_f/ \rangle \in \steps{*}{\auto}(t)$ if and only if
$\langle q'_f/ \rangle \in \steps{*}{\auto'}(t')$, where $q_f \in Q_f,\: q'_f
\in Q'_f$.

\begin{proof}
\there Suppose $\langle q_f/ \rangle \in \steps{n}{\auto}(t)$ for some $q_f
\in Q_f$ and $n \in \mathbb{N}$.  By Lemma~\ref{stmt:inst-of-steps}, there is
$v' \in \steps{n}{\auto'}(t')$ such that $\langle q_f/ \rangle \inst
\Gamma(v')$.  Therefore, we can say that $\Gamma(v') = \langle q'/ \rangle$,
where $q' \ni q_f$.  By definition of $\Gamma$, in this case $v' =
\Gamma(v')$.  Now, $q' \cap Q_f$ is non-empty (contains $q_f$), which means
that $q' \in Q'_f$.

\back Suppose $v' = \langle q'_f/ \rangle \in \steps{n}{\auto'}(t')$ for some
$q'_f \in Q'_f$ and $n \in \mathbb{N}$.  Again, $\Gamma(v') = v'$.  Now,
$q'_f$ being in $Q'_f$ implies that $q'_f \cap Q_f$ is non-empty; that is,
there is $q$ such that $q \in q'_f$ and $q \in Q_f$.  Let $v = \langle q/
\rangle$ -- a tree from $T(Q \cup \{/\})$.  Since $q \in q'_f$, we can see
that $v \inst v' = \Gamma(v')$.  Then, by
Lemma~\ref{stmt:inst-of-steps}, $v \in \steps{n}{\auto}(t)$; and, as we
already know, $q \in Q_f$.
\end{proof}

\begin{corollary}
\label{stmt:complete}
For any FSTA there exists an equivalent complete FSTA, i.e.\ such that any
pair of states has a matching rule in $\rules$.
\end{corollary}

\begin{proof}
The automaton $\auto'$ built in the above proof is complete.
\end{proof}

% vim:tw=78:keywordprg=dict

\bibliographystyle{unsrt}
\bibliography{../refs}

\end{document}